# Exploring the Technology Landscape through Topic Modeling, Expert Involvement, and Reinforcement Learning

Ali Nazari[1]*, Michael Weiss[2]

[1] Sprott School of Business, Carleton University, Ottawa, ON, Canada
[2] Technology Innovation Management (TIM) program, Carleton University, Ottawa, ON, Canada

**Abstract**

In today's rapidly evolving technological landscape, organizations face the challenge of integrating external insights into their decision-making processes to stay competitive. To address this issue, this study proposes a method that combines topic modeling, expert knowledge inputs, and reinforcement learning (RL) to enhance the detection of technological changes. The method has four main steps: (1) Build a relevant topic model, starting with textual data like documents and reports, to find key themes. (2) Create aspect-based topic models. Experts use curated keywords to build models that showcase key domain-specific aspects. (3) Iterative analysis and RL driven refinement: We examine metrics such as topic magnitude, similarity, entropy shifts, and how models change over time. We optimize topic selection with reinforcement learning (RL). Our reward function balances the diversity and similarity of the topics. (4) Synthesis and operational integration: Each iteration provides insights. In the final phase, the experts check these insights and reach new conclusions. These conclusions are designed for use in the firm's operational processes. The application is tested by forecasting trends in quantum communication. Results demonstrate the method's effectiveness in identifying, ranking, and tracking trends that align with expert input, providing a robust tool for exploring evolving technological landscapes. This research offers a scalable and adaptive solution for organizations to make informed strategic decisions in dynamic environments.

## 1. Introduction

Firms must adapt their processes to stay competitive amid rapid tech change. However, these efforts often fail due to poor or delayed resource allocation (March, 1991). Technological advances and competition greatly affect firm's performance and they must balance exploring growth with using their existing skills (Walrave et al., 2011). The literature on organizational learning emphasizes that lasting success depends on continuous learning (Agrawal et al., 2022) and requires a balance between exploration and exploitation (Floyd & Lane, 2000; Gupta et al., 2006). Thus, in today's tech-driven world, firms need to find new knowledge and opportunities and have to use their current strengths to stay competitive (O'Reilly & Tushman, 2011). A key challenge for organizations is to learn from the tech environment. They must balance this amid rapid technological advances, even with domain experts and consultants. It is critical to integrate external or implicit knowledge, especially in fast-evolving tech fields.

### 1.1. Organizational learning, the proposed method, and application

As Bogers et al. (2018) noted, firms increasingly rely on open innovation practices to use external knowledge and adapt to technological changes. By integrating insights from external sources such as domain experts, consultants, and research institutions, organizations can enhance their innovation performance (Laursen & Salter, 2006). However, the process of integrating external knowledge is not without its difficulties. Firms often struggle to effectively incorporate this knowledge into their decision-making processes, particularly in rapidly changing technological environments. This challenge is further intensified by the inherent tension between exploration (pursuing new opportunities) and exploitation (optimizing existing resources), which can lead to missed opportunities and suboptimal resource allocation (March, 1991; Walrave et al., 2011). Zahra & George, (2002) defined absorptive capacity that is key to understanding how firms learn from their external environments. Absorptive capacity is the ability to use external knowledge that helps organizations adapt better to new technologies. However, building this capacity requires both technical skills and a deep understanding of the changing tech landscape. For instance, Ansari & Garud, (2009) show how firms navigate intergenerational transitions in sociotechnical systems, like mobile communications, by learning and adapting to new technologies. Eggers & Park, (2018) also stress the need for firms to learn that helps them adapt to disruptive tech through experimentation and knowledge integration. We combine technical expertise and domain knowledge to aim to tackle the challenge of using external knowledge in strategic decision-making in the tech world (Lindgreen et al., 2021). Technical expertise helps organizations detect tech changes using data analytics, machine learning, or reinforcement learning. This support leads to quick trend identification and better decision-making (Agrawal et al., 2022; Blei et al., 2003; Sutton & Barto, 2018). Without understanding the domain, technical insights might miss the mark in real-world applications (Benner & Tushman, 2015). Domain knowledge keeps strategies on track. It helps interpret technical findings and balance exploration with exploitation. Plus, it makes the most of new technologies to gain a competitive edge (O'Reilly & Tushman, 2011; Zahra & George, 2002). We add an integration layer of both concepts to help organizations learn and adapt timely. It improves their balance between exploring new technology and using existing ones and boosts dynamic capabilities.

---

* Corresponding author. E-mail addresses: ali.nazari@carleton.ca

To address the learning challenges, firms must first organize their textual data such as internal and external reports and documents effectively and then analyze it by their domain experts. Topic modeling, for instance, enables firms to categorize and extract meaningful themes from large datasets, helping them identify emerging trends and technological shifts (Blei et al., 2003; Walrave et al., 2017). Once the data is organized, firms must fine-tune this information to align with their strategic processes. This requires domain experts to interpret the insights, understand the firm's current situation, and identify new opportunity for innovation and growth (O'Reilly & Tushman, 2011; Zahra & George, 2002). This iterative process must be ongoing to identify suitable advancements that are both practical and applicable. To organize and interpret data and involve experts, firms need a strong internal platform. This platform should enable experts to share knowledge and foster collaboration and consensus on strategic decisions (Laursen & Salter, 2006). Expert input is vital for refining their textual models. It ensures their relevance to real-world contexts and guides the search for new opportunities (Benner & Tushman, 2015). Besides, firms must use advanced techniques, like Neural Networks or Reinforcement Learning (RL), to simulate expert activities and explore new ideas (as stated in Sutton & Barto (2018). RL, for instance, helps firms make better predictions (Agrawal et al., 2022) and decisions by balancing exploration and exploitation. Exploration means seeking new opportunities, while exploitation focuses on using existing resources. This balance lets companies adapt to changes in their environment and keep up with technology (Eggers & Park, 2018).

*A proposed method*

This research proposes a method to improve organizational learning. It combines topic modeling, RL, and expert input. Topic modeling extracts and categorizes themes from large datasets that helps firms stay informed about tech advancements (Blei et al., 2003). Expert knowledge refines these insights. It ensures their accuracy and alignment with organizational goals (Walrave et al., 2017; Zahra & George, 2002). RL optimizes resource allocation by making balances exploration and exploitation activities. This lets firms adapt to new information and changing environments (Gupta et al., 2006; Sutton & Barto, 2018). Together, these components create a framework that improve learning and decision-making amid rapid tech changes.

*Application*

This integrated method provides several key benefits across different domains. In healthcare, firms can use the method to track advances in medical tech, like AI diagnostics and personalized medicine. They can then align their strategies. In finance, this approach helps spot trends in blockchain, cryptocurrency, and regulations. This way, firms can stay ahead of market changes and can also monitor advances in automation, 3D printing, and green manufacturing in manufacturing. Firms can monitor developments in automation, additive manufacturing, or sustainable production techniques, ensuring they remain competitive in a rapidly evolving landscape. By using topic modeling, expert input, and RL, organizations can improve their ability to respond to tech changes. This will boost their competitiveness and innovation (Eggers & Park, 2018; C. A. O'Reilly & Tushman, 2011). In quantum communication is advancing quickly and has major security implications and technologies, like quantum key distribution (QKD) and quantum repeaters, promise to revolutionize cryptography and data security. They enable unbreakable encryption (Bennett & Brassard, 2014; Liao et al., 2017). However, these advancements threaten existing encryption methods. Quantum computing could break traditional cryptographic protocols (Gisin et al., 2002; Hassija et al., 2020). Applying the integrated approach to quantum communication highlights its utility. Firms can use topic modeling to analyze research papers, patents, and industry reports. It can identify trends in QKD protocols or quantum-safe cryptography. Expert input makes sure these insights are relevant. RL guides the exploration and prioritizes areas that align with our goals and tech advancements (Cavaliere et al., 2020; Manzalini, 2020). This lets organizations update their cryptographic strategies. They can tackle new threats and use new opportunities. In high-tech fields like quantum communication and computing, the need for learning is even greater. Manzalini (2020) discusses that quantum communication technologies pose challenges and opportunities for firms. These technologies are evolving quickly. Organizations must constantly update their knowledge and strategies to stay ahead. Hassija et al. (2020) further underscore the implications of quantum computing for organizational adaptation, particularly in addressing security and innovation challenges. Firms must learn about these new technologies. They must also integrate them into their existing systems and processes.

The findings of this research are not limited to a specific industry or technological domain; rather, they offer a generalizable framework for enhancing organizational learning and decision-making in the face of rapid technological changes. By integrating topic modeling, reinforcement learning (RL), and expert involvement, the proposed method provides a versatile approach that can be adapted to various contexts.

## 2. Related work

Topic modeling, reinforcement learning (RL), and expert-in-the-loop systems have evolved together. This evolution has paved the way for adaptive knowledge discovery. This section brings these ideas together and highlights gaps in dynamic adaptability and the new capabilities that come from combining RL with expert feedback.

Topic modeling started with Latent Dirichlet Allocation (LDA) (Blei et al., 2003). This method uses probability to find hidden topics. It does this by modeling how documents relate to these topics. Extensions such as Correlated Topic Models (Blei & Lafferty, 2007) added topic correlations. Supervised versions (Mcauliffe & Blei, 2007) included metadata. These methods improved, but they still had trouble capturing nonlinear semantic patterns. They also lacked ways for ongoing refinement. Neural topic models, like the Neural Autoregressive Topic Model (Larochelle & Lauly, 2012), use deep learning to enhance topic coherence. Deep generative methods, such as Variational Autoencoders (VAEs) (Srivastava & Sutton, 2017; Xu & Durrett, 2018), also help with scalability. However, (Dieng et al., 2020) point out that even the best neural models often create static representations. They struggle to adapt to changing data and do not include domain expertise.

RL is a strong technique for making decisions step-by-step in changing situations (Agrawal et al., 2022; Sutton & Barto, 2018). Recent work bridges RL and topic modeling to address temporal adaptability. Foundational theories like adaptive dynamic programming (ADP) show how it improves feedback control. This happens through iterative reward signals (Lewis et al., 2012; Lewis & Vrabie, 2009). Recent work uses RL for topic modeling in changing environments. This applies to the principles of knowledge discovery. Agrawal et al., (2022) showed that RL agents can improve topic discovery. They use reward signals based on semantic coherence metrics. This helps models adapt to changing data distributions. Miao et al., (2016) built on earlier work in neural topic modeling. They pioneered neural variational inference for text processing. This created a framework for scalable probabilistic topic discovery. Later, RL-based methods expanded on their work.

Automated topic modeling has improved a lot, but domain expertise is still crucial. It helps make sense of the results and connects them to real-world uses. Early efforts included experts like Knowledge-Aware Bayesian Deep Topic Models (Wang et al., 2020). They added predefined domain ontologies into neural models. However, such approaches lack bidirectional interaction; experts could not iteratively refine models' post-deployment. Recent frameworks, such as tBERT (Peinelt et al., 2020), mix BERT embeddings with topic modeling. This method aimed to enhance semantic alignment, but it depended on fixed expert annotations. RL-enabled systems use expert feedback as a reward function. This helps models change their topic distributions based on user input. An example is context-guided embedding adaptation (Chen et al., 2014). This skill is especially useful in low-resource fields, like emerging technologies. Experts help focus exploration on important topics that data-driven models may miss. Human-in-the-loop (HITL) strategies help close this gap. They add expert feedback right into machine learning workflows. (Wu & Wang, 2024) show that human annotation helps improve tasks like topic modeling and text summarization. Also, (Monarch, 2021) outlines HITL frameworks for active learning. These frameworks balance automation with domain expertise. State-of-the-art methods, such as those by (Mosqueira-Rey et al., 2022), optimize iterative learning loops to ensure models adapt to user intent in evolving environments. These advancements work together. They create systems that combine automated discovery with expert guidance. This helps adapt and overcome challenges. For example, they tackle issues like concept drift and domain alignment, especially in low-resource settings.

Despite progress, key challenges persist. Scalable RL training for topic models is still very demanding on resources (Agrawal et al., 2022). Also, real-time adaptation struggles due to delays in expert feedback loops (Zhao et al., 2021). Models like tBERT (Peinelt et al., 2020) have a hard time balancing detailed meanings and themes. Recent advances in meta-RL and human-in-the-loop RL suggest solutions. They pre-train agents on varied datasets and improve how feedback is integrated. These methods highlight the benefits of hybrid systems. They combine RL's flexibility with expert guidance. This blend offers advantages not found in purely data-driven or static expert models. Previous systems, like probabilistic, neural, or RL-based ones, do not work well with experts. They miss out on closed-loop collaboration and iterative feedback. Our framework solves this by using expert feedback as a dynamic reward in an RL pipeline. This allows the system to: (1) focus on topics users choose in real time, (2) adapt to changes without retraining, and (3) tackle cold-start issues in new areas.

Quantum Communication Research

Quantum cryptography has advanced greatly in the past decades. Key theories and practical breakthroughs have changed secure communications. The 1980s saw the rise of quantum key distribution (QKD). In particular, the BB84 protocol by (Bennett & Brassard, 2014) was a major milestone in secure key exchange. Theoretical work on cryptography advanced in the 1990s. It included using quantum states for secure communication. Researchers, like (Cavaliere et al., 2020), explored QKD protocols. They studied their integration with classical cryptographic systems. This work underscored cryptography's evolving role in secure communications.

In the early 2000s, experiments tested QKD's practical uses. They included its integration into tele-communications infrastructure (Manzalini, 2020). A major achievement was deploying QKD over fiber-optic networks. It was the commercialization of QKD systems for secure transactions (Gisin et al., 2002). In the 2010s, satellite-based quantum cryptography became a focus. Liao et al. (2017) showed QKD between satellites and ground stations. This work extended the reach of secure communication networks.

Quantum cryptography is secure. It resists attacks by quantum computers using (Shor, 1994) algorithm. This is vital for protecting data against new threats. (Hassija et al., 2020) examined the role of QKD in safeguarding information against quantum computing risks. There is growing interest in combining cryptography with classical systems. Hybrid solutions aim to improve security in banking, healthcare, and government (Cavaliere et al., 2020). The field is working to solve scalability issues for wider use. Key research efforts focus on improving repeaters and photon detectors. These aim to extend the range of secure communication. Meanwhile, efforts to develop quantum-resistant encryption standards are underway. They seek to counter new threats from quantum computing (Manzalini, 2020; P. D. O'Reilly et al., 2022). Milestones like the BB84 protocol (Bennett & Brassard, 2014) laid key principles. Practical advances, like satellite-based quantum key distribution (QKD), pushed secure communication limits. These innovations are key to a strong quantum communication network. They are vital in an age shaped by quantum tech.

The limitations identified in the methods reviewed in the literature underscore significant challenges. Researchers have made advancements in topic modeling using deep learning. Yet, their complexity and high demands hinder real-time apps (Peinelt et al., 2020; Sutton & Barto, 2018). Newer models using expert feedback (Wang et al., 2020) fix these issues. However, they still struggle with scalability. Also, they need to use domain-specific knowledge to find adoptable technologies. Besides, RL shows promise for dynamic knowledge discovery. Yet, real-time apps in large data repositories are still evolving (Sutton & Barto, 2018; Yen et al., 2002). Quantum cryptography domain faces these barriers to adoption (Cavaliere et al., 2020; Manzalini, 2020). These issues show a need for more innovation in both theory and practice.

## 3. Method

The proposed method integrates topic modeling and RL techniques with a structured framework for data analysis. It begins with an unsupervised learning algorithm, LDA from (Blei et al., 2003), enhanced by expert inputs to generate aspect-specific topic models. It also uses RL capability, focusing Q-learning technique (Sutton & Barto, 2018; Yen et al., 2002) to creates a cycle of continuous improvement. We adopt an exploratory approach to test our method. This data serves as a foundation for uncovering emerging technologies and identifying advancements. To ensure robustness, we analysis the findings from the method. It encompasses two primary objectives: 1. Modeling Screened Data: To uncover latent patterns and structures. 2. Analyzing Technological Shifts: Extracting novel insights.

To achieve the objectives, the method involves an iterative process. It has several key steps: First, we collect domain-specific documents through targeted search keywords in online libraries, forming a corpus. Next, a topic modeling technique is applied to the corpus to create an initial topic model. Aspect-based topic models are created by weighting keywords in the original model's topic word distributions. Experts incorporate domain knowledge to refine these models, ensuring their relevance and accuracy. To simulate expert refinement, we bring in external sources, such as conference papers. This allows us to access domain expertise that goes beyond the corpus. These sources reflect emerging trends and priorities in the field, serving as a dynamic "expert signal" to guide topic model updates. RL algorithm is then applied as an iterative process. It refines topic selection and analysis to improve over time. Finally, an evaluation analyzes the results to identify key insights and trends. This process promotes teamwork between machines and humans. As we discussed in review section, it creates feedback loops that improve the analysis.

To clarify the process, we present a pseudocode representation of our method (Table 1). Phase 1, 'Data Collection,' condenses the topic modeling process. In our earlier paper (Nazari & Weiss, 2025), we elaborate on the details of Phase 1 and portions of Phase 2, Lines 1 to 6 in Table 1. These phases outline steps including collecting papers, defining keywords, organizing corpora, generating baseline topic models, and developing aspect-based models with expert keywords.

**Table 1: The proposed overall algorithm[3]**

| Phases/Steps //Step description |
|---|
| *Phase 1: Data Collection* |
| 1: SK ← define_search_keywords (domain) // *Define search keywords specific to the research domain with input from domain experts.* |
| 2: C ← build_corpus (SK) // *Collect documents using defined keywords, followed by text preprocessing to create a refined corpus.* |
| *Phase 2: Topic Modeling and RL (Topic Model Analysis)* |
| *Topic Model Analysis (Initial Steps)* |
| 3: TM ← LDA(C) // *Apply LDA to the corpus to generate the initial topic model.* |
| 4: CTP1 ← Initial Topic Model (TM) // *Use the generated topic model as the starting point (CTP1).* |
|     CTP2 ← None // *only in first iteration* |
| |
| *Aspect-Based Refinement (Iterative Steps)* |
| 5: AText ← Aspect_Identification (Domain Expert notes) // *Identify aspect-related keywords based on expert input; in this research, we gather aspect-related keywords using this method: 1-* **Text Collection:** *Collect text from targeted resources like reports, conferences, or workshops. 2-* **Preprocessing**: *Clean text using tokenization, stop-word removal, and lemmatization. 3-* **Keyword Extraction**: *Apply TF-IDF to rank terms by importance within the conference corpus. 4- Keep important terms with high* **TF-IDF scores**. *Focus on the top 100 of these terms. Also, remove general terms like "algorithm" and "application."; Papers → Preprocessing → Cleaned text → TF-IDF →* {"secure protocol" (0.89), "entanglement distribution" (0.76)}. |
| |
| 6: AT ← Weighted_Aspect_Keywords (AText) // *Assign weights to the keywords based on TF-IDF scores. Normalize weights to emphasize high-priority terms.* |
| |
| 7: ATM ← get_AspectTM (CTP1, AT) // *Refine the initial topic model (CTP1) by incorporating the aspect-based keywords to generate the Aspect Topic Model (ATM).* |
| 8: CTP2 ← ATM // *Assign the refined model as CTP2 for comparison with CTP1.* |
| *RL Process* |
| 9: current_state ← compare_models(CTP1, CTP2) // *Create topic similarity, magnitude, entropy changes and topic absolute difference between CTP1 and CTP2 to calculate future rewards for finding suitable action.* |
| 10: action ← find_topics (current_state) // *Select an action (topic(s)) based on approximate future rewards of topics.* |
| 11: new_state ← adjust_topic_model_with_new_state (action, CTP2, new_keywords, new_documents) |
| // *Transition to a new state by incorporating new keywords and documents.* |
| 12: reward ← calculate_reward(new_state, action) // *Calculate topics reward with new related documents to validate the selected topics.* |
| 13: RL_Model ← update_RL_model(action, reward) // *Update the RL model's policies and hyperparameters based on the calculated rewards and selected topics.* |
| |
| *Phase 3: Analysis* |
| 14: VR ← compare_topic_models(CTP1, CTP2) // *Conduct heatmap analysis to compare changes and alignments in CTP1 and CTP2.* |
| 15: Patterns_Novelty ← Technology_Vision (QCrypt23 or 24 with CTP2_Allwords) // *Analyze the selected topics and their alignment with new documents by examining their associations to uncover novel insights or patterns in the domain.* |

---

[3] Q-learning Topic Selection for Topic Modeling Script

16:  fine_tuned_topics = fine_tune_topics (CTP1&2, DocsCTP2, Patterns_Novelty)  *// Further refine topics based on insights from analysis and patterns in DocsCTP2 or DocsCTP3; evaluate the results and selection process with examine word trends and documents association with the selected topic(s) in each iteration.*

*Iteration Transition*
17: CTP1 ← CTP2                                *// Replace CTP1 with the previous iteration's CTP2 to prepare for the next iteration.*
18: CTP2 ← new_state                    *Assign the current refined model to CTP2 for further iteration.*

If (Patterns_Novelty)                        *// Stop the process if significant novel patterns are identified. Otherwise, continue*
   End Episode                        *with the iterative process by reinitiating from Step 5.*
Else
   Proceed to Step 5

Later steps in the second phase use RL algorithm to refine topics through ongoing adjustments. The primary focus is on RL for the iterative refinement and adaptation of topic models. The process starts by comparing the initial topic model (CTP1) with an aspect-based model (CTP2). In our context, CTP stands for Contextual Topic Perspectives. 'Contextual' refers to the different viewpoints for analyzing the topic model. 'Topic Perspective' means that each model version offers a unique view of the topics. It shows how the model evolves or is re-evaluated over time from different views. We seek significant differences, similarity, and entropy changes within topics based on a predefined policy. RL selects action to adjust the topic model. It does this by incorporating new keywords and documents. In each iteration, the agent calculates approximate rewards to calculate topic Q-values in q-learning RL algorithm. The novelty and improvements of the topic model form the basis. The RL model then updates to inform future actions. This iterative process allows for continuous model enhancement by integrating new, relevant information. We verify the model's effectiveness by analyzing the results in phase 3. This includes heatmaps and cosine similarity and rewards comparisons. We look for new insights and topic evolution and also ensure the RL-driven updates align with technological advancements.

Phase 1: Data Collection

We define search keywords from key papers of the predefined domain of research. These keywords guide the collection of domain-specific corpora from online sources, like Web of Science and Scopus. This data forms the basis for constructing a corpus. Step 1 is to identify and define domain search keywords (SK) from a set of domain-related documents, like peer-reviewed articles, reports, and patents. After retrieving the corpora, we build a refined corpus[4] (C) in Step 2. We do this by applying some text-processing techniques. This includes stop word removal, stemming, lemmatization, and tokenization. It also includes steps to ensure the corpus is clean and structured and uses a neural network technique to screen out irrelevant documents. It must be ready for the next phase: topic modeling. These steps are vital and ensure the data's quality and relevance for further analysis.

Phase 2: Topic Modeling and RL

This phase begins with using LDA algorithm (Blei et al., 2003) on the preprocessed corpus (Step 3). Researchers use methods like variational inference (Blei & Jordan, 2006) or Gibbs sampling (Griffiths & Steyvers, 2004) to do this. This will create an initial topic model (TM). The model represents the distribution of words and documents across topics. LDA aims to find the posterior distribution using variational inference. It does this with the latent variables (topics) and the observed data (documents and words). Step 4 establishes the starting point for the RL process. The system initializes the state CTP1 as the initial topic model (TM). It is the baseline model before any updates. Meanwhile, we set CTP2 to 'None' to state that no one has yet prepared an aspect-based topic model. This setup allows the algorithm to start an iterative process. It will compare and refine topic models using new data and defined policies.

Next, we (Step 5 & 6) identify aspects (AText in Table 1) using an agent-based aspect identification approach. Agent-based aspect identification is an automated process. A software agent, which is a rule-driven NLP pipeline, extracts and prioritizes keywords from domain related texts. The approach performs the following tasks: 1- Automatically gathers domain related papers, like the QCrypt 2023 proceedings, from set sources such as arXiv and conference sites (Text Retrieval) and then cleans text using tokenization, stop-word removal, and lemmatization (Preprocessing). 2- Applies TF-IDF to rank terms by importance within the conference corpus (Keyword Extraction). It uses thresholding, like keeping terms with TF-IDF scores over 0.7. It also has exclusion rules, such as removing generic terms like "analysis" or "method" or irrelevant previous aspect keywords. This helps create a list of candidate keywords (Aspect Filtering). 4- Compare terms with a domain-specific ontology, such as the communication glossary, to ensure they are relevant (Aspect Validation) (AT in Table 1). The aspect identification agent automates keyword extraction. It uses rule-based NLP techniques, like TF-IDF scoring and thresholding. This helps rank new terms from sources such as conference proceedings. While the agent preprocesses, ranks terms, and filters, it uses human knowledge in three ways: 1- It sets TF-IDF thresholds and exclusion lists to highlight important terms. 2- It creates ontologies showing relationships, such as "post-quantum cryptography" under "security protocols." 3- It checks results to remove irrelevant options. This hybrid method combines automation for scalability and expert rules for accuracy. It ensures that extracted elements fit domain needs and reduces manual work. In our research, we take input from related conference papers to extract keywords. You can see the top keywords in Figures 2 and 7.

We also use the aspect's weighted keywords (AT) and the initial topic model (TM) to train the TM and compute relevance scores for each topic (Step 7). We do this by multiplying the aspect keywords vector with the topic keywords vector, focusing only on the intersection of the two vectors. This gives us what we call aspect-topic models (ATM). The scores show how relevant documents of each topic are to the identified aspects. It assigns different weights to the topics based on aspect keywords. We have outlined these steps taken thus far in (Nazari & Weiss, 2025). These two models, CTP1 and CTP2, serve as inputs to the RL process (Step 7 & 8). CTP1 is the initial topic model and CTP2 is the aspect-based topic model. In RL process, an agent identifies topics likely to contain new information for further analysis. It operates based on predefined policies, hyperparameters, and thresholds. Domain experts, later in end of each iteration, will then examine the insights derived from these novel topics.

RL Process

This stage employs reinforcement learning (RL) to improve topic models. It balances two key activities: exploring new topics and enhancing coherence. We set up topic selection as a Markov decision process. A Q-learning agent improves actions (topic adjustments) over time (Bishop, 2006; Sutton & Barto, 2018). It does this by using approximate rewards from four metrics: 1- Topic Magnitude: This measures how different the distributions are between the baseline (CTP1) and the refined model (CTP2). It helps highlight new themes. 2- Cosine Similarity: Quantifies topic-document alignment to maintain coherence. 3- Entropy Changes: Tracks uncertainty shifts to guide topic specialization. 4- Absolute Difference (ADNS): Assesses normalized topic weight variations.

$$\text{Magnitude} = \sqrt{\sum (Topic\ Weights)^2} \quad (1)$$

The formula 1 calculates the magnitude of a topic weight vector by summing the squares of the topic weights and taking the square root. It measures the spread or intensity of topic contributions in a document.

Cosine similarity compares two topic word vectors, regardless of their size. It assesses their overlap (Manning & Schutze, 1999).

$$\text{Cosine Similarity} = \frac{A \cdot B}{\|A\| \cdot \|B\|} = \frac{\sum_{i=1}^{n} A_i * B_i}{\sqrt{\sum_{i=1}^{n} A_i^2} * \sqrt{\sum_{i=1}^{n} B_i^2}} \quad (2)$$

where A·B is the dot product, and ‖A‖, ‖B‖ are norms, vector magnitudes. CTP1 (initial cryptography topics) and CTP2 (cryptography-protocols) are compared to examine topic evolution. The similarity matrix helps track consistency and topic shifts. Low similarity indicates significant changes, signaling redefinition or new focus. It detects how topics evolve and guides adjustments (Blei & Lafferty, 2007).

Entropy measures uncertainty in topic distributions. It shows if topics become more specialized or broad (Shannon, 2001). High entropy suggests a broad, unclear topic, while low entropy shows a more focused one (Blei & Lafferty, 2007). Tracking entropy changes helps identify whether topics should be refined or explored further:

$$H(T) = -\sum_{i=1}^{n} P(w_i|T) \log P(w_i|T) \quad (3)$$

Entropy H(*T*) quantifies the uncertainty or diversity in the word distribution of a topic *T*. Higher entropy reflects a more uniform distribution, indicating less focus on specific words, while lower entropy suggests a more concentrated topic. It is calculated as formula (3), where $P(w_i|T)$ is the conditional probability of word $w_i$ in topic *T*, and $\log P(w_i|T)$ moderates contributions based on probability values. Entropy changes in two sequential topic models can signal topic evolution. Besides, using entropy changes to guide RL policy decisions to make an action on topic selection is effective (Xin et al., 2020). Increased entropy suggests diversification and decreased entropy implies a greater focus (Blei et al., 2003).

In addition, Absolute Difference in Normalized Sums (ADNS) compares two normalized vectors. It measures shifts in their distributions.

$$V_{norm} = \left[\frac{v_1}{\sum_{i=1}^{n} v_i}, \frac{v_2}{\sum_{i=1}^{n} v_i}, \ldots, \frac{v_n}{\sum_{i=1}^{n} v_i}\right] \quad (4)$$

$$\text{ADNS} = \sum_{i=1}^{n} \left|\frac{v_{1,i}}{\sum_{i=1}^{n} v_{1,i}} - \frac{v_{2,i}}{\sum_{i=1}^{n} v_{2,i}}\right| \quad (5)$$

The formula for this calculation usually involves two steps: First, we normalize each vector (distribution) by dividing each element by the sum of all elements in the vector. This ensures that the sum of all elements in the vector equals 1, transforming each vector into a probability distribution. For a vector $V = [v_1, v_2, \ldots, v_n]$, the normalized vector $V_{norm}$. Second, after normalization, the absolute difference is calculated between the sums of corresponding elements in two normalized vectors $V_1$ and $V_2$. If $V_1$ and $V_2$ are normalized versions of two topic distributions. Where $v_{1,i}$ and $v_{2,i}$ are the elements of the vectors $V_1$ and $V_2$, and $\sum_{i=1}^{n} v_{1,i}$ and $\sum_{i=1}^{n} v_{2,i}$ are the sums of the elements in $V_1$ and $V_2$, respectively. This formula calculates the total difference between the two normalized distributions. It is a simple method to compare the divergence of two distributions after normalizing their elements (Blei & Lafferty, 2007; Kullback & Leibler, 1951).

Combining these metrics, the RL agent calculates approximate rewards. For a balanced evaluation of topic model updates, we combine magnitude and entropy as novelty metrics to measure the introduction of new patterns, with cosine similarity and ADNS as stability metrics to ensure topic coherence. The reward function combines four key metrics: magnitude, similarity, entropy, and ADNS, weighted by coefficients (λ1, λ2, λ3, and λ4). This prioritizes exploration or stability, fostering adaptation (Ng et al., 1999), and helps the RL agent make decisions that align with our learning goals. Magnitude shows how strong topic contributions are. It helps organizations detect and rank new trends, like secure authentication. Similarity keeps us aligned with what we know. It helps us make consistent decisions. This way, we avoid major changes that move away from our main skills. Entropy shows topic uncertainty. It balances two important areas. First, it explores new ideas, like quantum networks. Then, it focuses on practical applications that boost advancements, such as entanglement swapping. ADNS detects changes in topic distributions. This signal shifts in paradigms that need new strategies. For example, it helps in adapting to emerging post-quantum cryptographic protocols. The weighted sum of these components helps rank tasks dynamically. Organizations can adjust λ1−λ4 to focus on different goals. They can detect trends, ensure stability, drive innovation, or adapt as needed.

$$\text{Approximate Reward } (R(s,a)) = \lambda1 \times \text{Magnitude} + \lambda2 \times \text{Similarity} + \lambda3 \times \text{Entropy} + \lambda4 \times \text{ADNS} \quad (6)$$

By tuning the metrics' coefficients, we can adjust the reward function. This design lets the RL agent explore many paths for the topic model to promotes innovation or refinement. In our test process, we assign greater weight to magnitude and less to similarity to effectively explore new aspects of topics and currently keep the other metrics low. The coefficients are λ1=0.75 for magnitude, λ2=0.15 for cosine similarity, λ3=0.05 and λ4=0.05 for entropy changes and ADNS respectively. For early iterations, higher λ1 most probably can encourage exploration, while later iterations increasing λ2 facilitates exploitation the new data. We test different coefficients to examine approximate rewards and maximum Q-values to select topic candidates. The RL agent calculates expected Q-values as outlined by Watkins & Dayan, (1992):

$$Q_{t+1}(s,a) = (1-\alpha) \times Q_t(s,a) + \alpha \times (R_t(s,a) + \gamma \times max a' Q_t(s',a')) \quad (7)$$

where $Q(s,a)$ is the current value of the state-action pair of aspect topic model and selected topics (state *s*, action *a*). Weights of topics in CTP2 refining with expert-driven aspect keywords and the action is selecting and refining topics with greater Q-values calculated using approximate rewards. $\alpha$ is the learning rate which determines how much the new reward influences the previous Q-value. $R(s,a)$ is the approximate reward (as defined in Formula 6) for taking action *a* in state *s*. It is based on the average weight of topics after applying new keywords (reflecting improved topic relevance and novelty). $\gamma$ is the discount factor which balances the value of immediate vs. future rewards. $max a' Q(s',a')$ is the maximum expected future reward for the next state *s'*, considering the best action *a'*. For our model, this refers to how the RL agent would evaluate future topic refinements in CTP2, considering which actions (e.g., further keyword adjustments or topic selections) would lead to the best future topic improvements as shown by (Mnih et al., 2015). It ensures that the agent prioritizes actions that will lead to the most promising improvements in future topic models. According to the pseudocode, the RL steps as follows:

Step 9: CTP2 is updated with protocol keywords and specific protocol advancements (Steps 4 & 8). The comparison of these two models, in each iteration, defines the system's current state. Metrics like topic similarity, magnitude, entropy changes, and topic absolute difference are calculated. They find approximate future rewards and guide action selection.

Step 10: Next, we evaluate the similarity between topics in CTP1 and CTP2, alongside the magnitude of topics. We also examine the entropy changes in the transmission. They measure changes in topic distributions based on the weighted aspect keywords. To select topics, the agent approximates the topic reward first. We consider topic magnitude, similarity scores, entropy variations, and ADNS to calculate it (Sutton & Barto, 2018). And then, the Q-values are calculated based on Formula (7). The maximum Q-values based on the approximate rewards are selected for future investigation and analysis. We select the top five topics for further analysis.

Step 11: Once the agent selects topic(s), an action, it adjusts the topic models. This process adds documents from the other sources like papers from 2023 and 2024 conferences to evaluate the selection process. The initial topic model is created on data up to 2022. We include the following years to show the effectiveness of the topic selection process. The agent uses the action to infer new topics from the current CTP2 model. It then transforms into a new state. This transition introduces new topic models. They reflect the changing state of the domain.

In Step 12, to assess whether the selected topics align with the technology trends desired by the expert, we adapt the reward function in our RL framework based on the explicable reward design approach by Devidze et al., (2021). Their method aims to align reward functions with clear, context-specific criteria. It ensures the agent's learning aligns with both set goals and new patterns. The updated Q-values, based on these rewards, show how the selection process aligns with market technologies and the expert's goals. We use a modified reward function. It calculates expected rewards using the 2023 and 2024 conference papers. It also boosts the exploration rate by applying changes in entropy. For our case, it emphasizes trends in post-quantum cryptography and secure protocols. It guides the agent to recognize clear progress in these evolving topics.

Modifying Rewards in the Q-Learning Formula for Enhanced Exploration

To bring up the exploration rate in our small topic model, we add entropy changes to the reward function as well. The formula for the modified reward function is as follows:

$$R(s,a) = R(s,a)_{base} + \lambda \cdot Entropy(s,a) \quad (8)$$

The formula represents the total reward for selecting action $a$ (topic) in state $s$. The system derives the base reward, denoted as, from the average cosine similarity between each CTP2 topic and the new documents. Less similarity results in a high base reward. This calculation and document can be refined in each iteration with input from multiple experts, ensuring all experts follow a unified scenario to reach a consensus on adopting the emerging technology. We calculate this using:

$$R(s,a)_{base} = \frac{1}{d} \sum_1^d (Cosine\ Similarity(T_j, D_i) > t) \quad (9)$$

Here, d is the number of new documents (e.g., the 2023 conference papers). We consider these documents are expert input. The cosine similarity measures the alignment of each topic with the documents. We also consider t as a threshold of the similarity scores to get the sum of the most associated documents to the topic $a$ (e.g., 0.3). In formula 8, $Entropy(s,a)$ represents the degree of uncertainty or novelty in the topic, derived from its word distributions in CTP2. A higher entropy value indicates that the topic is more diverse or underexplored. The hyperparameter λ controls the weight given to the entropy. It helps balance exploring new topics with refining existing ones.

Step 13: The system updates the RL model based on the reward received. If an action leads to little or no reward, the system learns to avoid similar actions or policy. For example, introducing redundant keywords that do not improve the model. This step reinforces successful actions that get a higher reward and higher Q-values. It helps the system select actions for improved topic modeling with greater efficiency. Over time, the RL model becomes more adept at identifying and incorporating new, valuable topics as it processes evolving data. We update the Q-values (based on formula 7) and save the parameters (α, γ, and λ). Actions that lead to high rewards have their Q-values increased, making them more likely to be selected in the future. We discourage actions with low rewards as their Q-values decrease. With each run, the RL model learns which actions improve objectives, like topic relevance. These actions include adding keywords or modifying topics. Reinforced successful actions help the model. It can now find new topics, avoid harmful actions, and process changing data with greater efficiency. For example, the model may first introduce keywords without a specific pattern. It then evaluates their impact based on coherence or expert feedback. It assigns higher rewards to meaningful keywords and lower rewards to irrelevant ones. Over time, the model prioritizes actions that consistently enhance topic quality.

Phase 3: Analysis

The analysis aims to find new insights and track topic changes. Step 14: Heatmap analysis compares CTP1 and CTP2 to assess the similarity between topics in different aspects. The heatmap shows how well these aspects align. It clearly shows any topic overlap or differences. Step 15: Use cosine similarity to compare documents with the topic matrix. It will show how closely each document aligns with selected topics. Higher cosine similarity values show stronger relevance. They help to identify shifts in focus within the topics. Step 16: Topics are refined by incorporating relevant new documents, enabling the model to evolve over time. This RL-driven update process improves the model. It can now spot trends and adapt to new data. This keeps it relevant and responsive.

The final step in the loop is preparing for the next iteration. The updated topic model CTP2 now becomes the new CTP1 for the next comparison. The inferred new state becomes the new CTP2 (Steps 17 and 18). The system compares, adjusts, and refines the models in response to the entry of new data and keywords. The loop continues running until the desired results are achieved in each episode. At the end of an episode, another path extension may begin. This is when either we reach a desired novel topic(s) or we can make no more significant improvements. The final result is a refined set of topic models. They evolved over many iterations, incorporating keyword changes and new documents. They now offer a more accurate view of the domain technology landscape. This process ensures the model learns and adapts. It will align with new trends while keeping a focus on the domain's core aspects.

## 4. Results and Analysis

We use the quantum technology landscape to test our method. We analyze how our method selects topics, particularly those encompassing security protocols that emerge through RL-driven topic selection. Our dataset is limited to papers we found in two libraries. We will examine keyword shifts within topics in the modified topic models influenced by aspect-weighted keywords. The subsections in the results align with the phases and iterative steps of our method. Phase 1 (Data Collection) sets the foundation. It defines search keywords and refines the corpus[4]. Phase 2 (Topic Model Analysis) uses this to establish the baseline topic model (CTP1[5]). We collect text[6] about specific aspects of protocols in a conference like Quantum Tech[7]. The iterative, expert-guided process of aspect-based refinement aligns with the development of aspect-specific topic models, like CTP2[8] (Protocols) and CTP3[9]. These models use weighted keywords. Then it drives comparisons (e.g., CTP1 & 2[10], CTP2 & 3[11]). These find less similar topics or entropy shifts. They inform action selection and refinements. Each iteration uses updated datasets (e.g., QCrypt2023_Papers[12], QCrypt2024_Papers[13]) to calculate expected rewards. It also tests how well documents map to topics (DocCTP2[14], DocCTP3[15]). The results show this alignment through detailed matrices, entropy calculations, and document mappings. They illustrate the evolving insights into cryptographic themes and protocols. We organized our results into the following subsections.

---

[4] Corpus [A small dataset of documents published by 2022 collected from online libraries and screened for relevance]

[5] CTP1 [Initial topic model in first iteration]

[6] Aspect's text [Aspect's text and their keywords]

[7] Quantum. Tech

[8] CTP2 [Protocol aspect topic model]

[9] CTP3 [Security aspect topic model]

[10] CTP1 & 2 [Similarity, ADNS, and entropy changes metrics for CTP1 & CTP2]

[11] CTP2 & 3 [Similarity, ADNS, and entropy changes metrics for CTP2 & CTP3]

[12] QCrypt2023 [A set of documents derived from conference papers is used to calculate the expected rewards in first iteration]

[13] QCrypt2024 [A set of documents derived from conference papers is used to calculate the expected rewards in second iteration]

[14] DocCTP2 [Similarity of CTP2 topics with QCrypt2023 documents]

[15] DocCTP3 [Similarity of CTP2 topics with QCrypt2024 documents]

Step 1- 4: Development of the Initial Aspect-Based Topic Model (CTP1)
To create an aspect topic model, we follow the steps in the method.

Step 1: Define Search Keywords (SK): The process begins by identifying a set of search keywords from key papers in quantum communication and computing such as (Cavaliere et al., 2020; Hassija et al., 2020; Manzalini, 2020). The extracted keywords include 'quantum AND (communication OR network) AND (development OR application OR experiment OR implement OR algorithm OR use) AND (feasibility OR future OR forecast OR trend OR progress).' Then, we search two online libraries, Scopus and Web of Science. We collect documents that match the search keywords and create a corpus.

Step 2: Text Preprocessing: After collecting the corpus, we preprocess it. This refines and standardizes the text for analysis. The preprocessing steps are to remove stop words and irrelevant terms, normalize the text (e.g., lowercase, stem, and lemmatize), and filter for documents with the defined keywords. An initial search found 3,527 documents (1,600 from Web of Science and 1,927 from Scopus). Relevance scoring narrowed it to 2,406 top papers from journals. A neural network classifier refined the selection. It identified 1,048 relevant documents. The output is a refined corpus[4] (C) that is ready for topic modeling.

Step 3: We create a baseline topic model from the refined corpus (C) using Latent Dirichlet Allocation (LDA). This step finds the latent topics in the corpus. It provides an initial topic distribution across the document set. Using LDA on 1,048 preprocessed documents, we identified six primary topics. Then it divided each primary topic into subtopics. This process yielded 39 final topics, labeled T01 to T39.

Step 4: The output of the LDA model serves as the Initial Topic Model (TM). This model represents the primary cryptography-related themes and distributions derived from the corpus. We name it CTP1[5] in this process. CTP1 forms the base for later versions. They use aspect-based tweaks and RL to explore subtopics. These include protocols and new advancements. This step ensures the initial state aligns with the broader method. It aims to capture key aspects of cryptography. By following these steps, CTP1 is the starting point for the iterative, RL process.

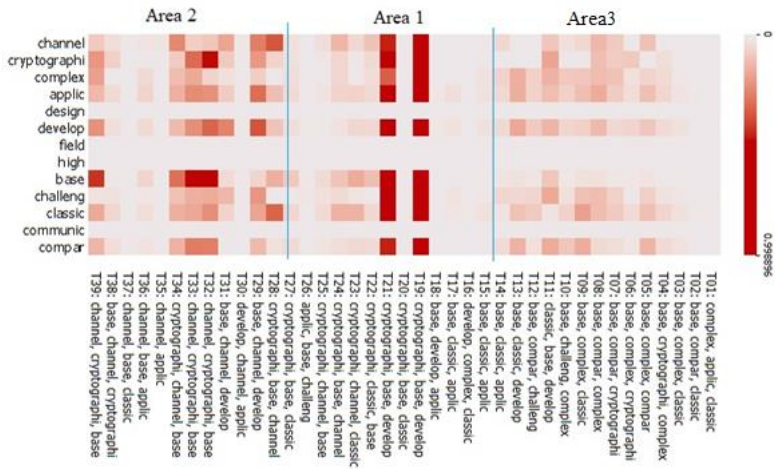

**Fig 1:** Word-Topic Distribution Heatmap: Top Words Across 39 Topics in CTP1

The CTP1 heatmap (Figure 1) shows word-topic associations. Color intensity reflects the strength of these relationships. The x-axis displays the topics with their top words derived from CTP1, while the model presents shared top keywords on the y-axis. The color gradient shows the strength of associations between words and topics. Darker shades show stronger relationships. The top keywords in CTP1 are complex, applic, classic, develop, challeng, base, compar, channel, and cryptographi. The heatmap shows distinct clusters of words. They are often linked to specific topics. For example, 'channel,' and 'cryptographi' are in the upper-left corner. They have a close connection to quantum key distribution topics, including T29, T31, T32, T33, T34, T35, T37, and T39. Some words appear in many topics, suggesting overlapping concepts or broader applicability. For example, researchers link 'learn' to several topics. Topics on machine learning or AI, like T1, T5, T6, T7, T9, and T11, focus on 'complex,' 'applic,' and 'develop.' Topics on cryptography and security, like T19 and T21, use 'challeng,' 'base,' and 'classic.'

## 4.1. ITERATION 1

**Itr1-Step 5 & 6: Domain Expert-Defined Weighted Keywords for Aspect 1**

We analyzed articles from Quantum Tech[7] to identify aspects using the agent-based method. We then summarized the documents into thematic categories. They represent current key aspects in the field. We focus on advances in cryptographic protocol aspect (labeled as Aspect 1[6]). For each aspect, we select 100 high-weighted keywords as aspect keywords. The top 10 for Aspect 1 are in Figure 3.

*Word Cloud of Aspect 1*

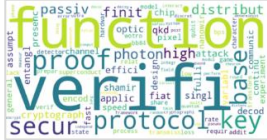

**Top keywords:** verifi-0.029, function-0.022, proof-0.022, protocol-0.02, secur-0.019, key-0.019, base-0.019, photon-0.018, distribut-0.017, high-0.017

**Fig 2:** Aspect keywords and word cloud: weighted keywords across aspect 1

The figure above shows a word cloud of the Aspect 1 text. It highlights the top keywords and their weights. It emphasizes 'verifi', 'function', 'proof', and 'protocol'. These keywords emphasize protocol verification, security, and key distribution in cryptographic communication systems. This selection guides the first iteration's analysis, prioritizing advancements in cryptographic protocols.

**Itr1-Step 7 & 8: Protocol Topic Model (Applying Aspect 1) (CTP2)**

After applying the aspect 1 keywords to CTP1, a broader set of keywords is now linked to the topics. The CTP heatmap's expanded keyword representation shows this. We try to keep the heatmap's structure consistent with CTP1. It shows that the core relationships between the majority of words and topics are stable. Yet, the intensity of the colors has shifted, reflecting changes in the strength of word-topic associations. Some words have developed varying degrees of association with particular topics. The heatmap displays words that have gained prominence through new weighting in specific topics. We present the combined words from CTP1 and CTP2 in Figure 2, maintaining three groups of topics to illustrate the shifts.

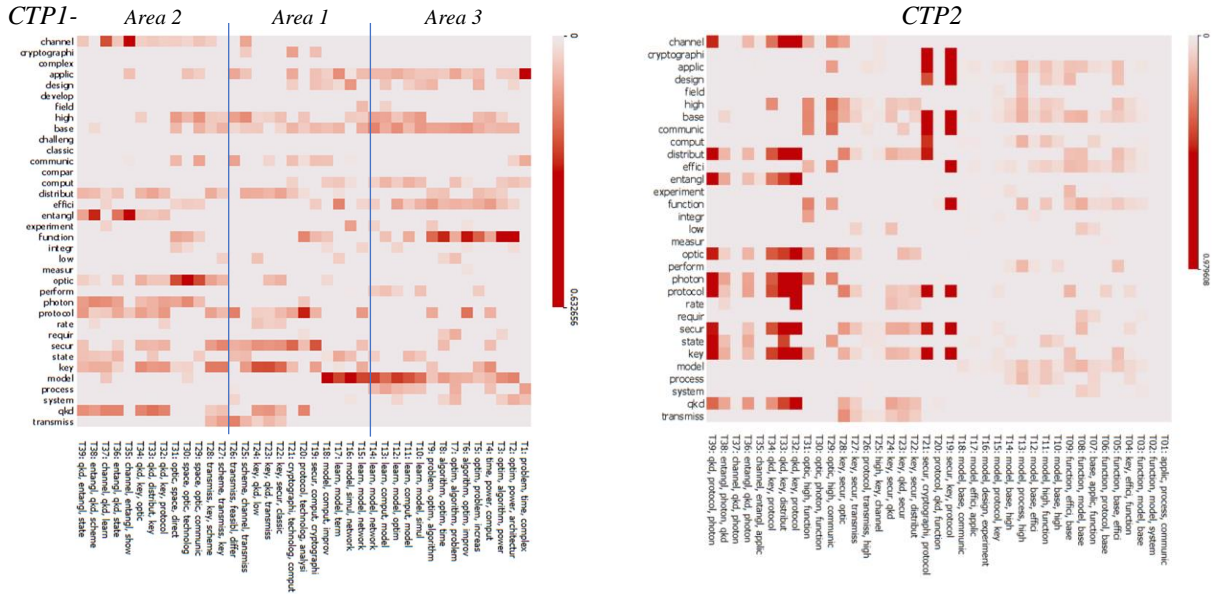

**Fig 3:** Word-Topic Distribution Heatmap: Top Words Across 39 Topics in CTP

As shown in the above CTP1 heatmap, there is a noticeable shift from areas 1 & 3 to area 2, across the topics from CTP1 to CTP2. Topics T19 and T21 have become less dominant. Keywords like 'model,' 'process,' 'function,' and 'applic' have gained prominence in area 3 in CTP1, especially in topics T1 to T18. Meanwhile, cryptography-related keywords like 'channel' and 'entangle' are in area 2. They span topics T22 to T39. The CTP2 heatmap shows that advancements in protocols are most linked to topics T22 and after. The word 'QKD' has a strong connection there. These topics likely focus on improving communication protocols. They involve security and transmission processes. Nearby terms like 'channel' (security), 'entangle,' and 'optic' suggest this.

Itr1-Step 9: Similarity Matrix Comparing CTP1 and CTP2 with Entropy Calculation for the RL Process (CTP1&2)

The heatmaps below show three matrices: 1. The similarity scores between topics in CTP1 and CTP2 (calculated using Formula 2). 2. The Absolute Difference in Normalized Sums (ADNS) between the word-topic vectors in CTP1 and CTP2 (calculated using Formula 1). 3. The entropy changes in topics in CTP2 (calculated using Formula 3). The greatest divergences involve in calculating Q-values finds topics that differ between CTP1 and CTP2. It focuses on those with the greatest Q-values for RL-driven refinements.

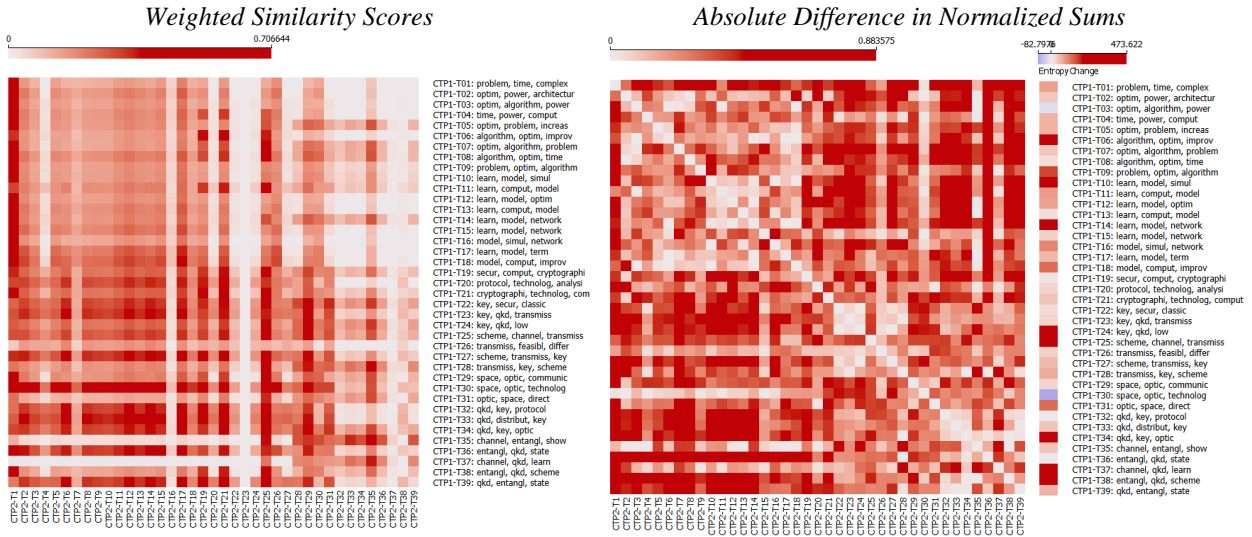

**Fig 4:** Matrices for Evaluating Topic Stability and Evolution in the First Iteration

The left heatmap compares topic distributions between two models, CTP1 and CTP2, as shown in CTP1&2 file[10]. CTP1 is the cryptography topic model (initial topic model). CTP2 is an updated version. Each row in the heatmap corresponds to a topic from CTP1, while each column represents a topic from CTP2, resulting in a 39x39 matrix. Matrix entries denote the similarity score between topics from the two models, with values ranging from 0 to 1. Higher values suggest stronger alignment. They show that topics have retained their identity across models. Lower values may reveal shifts in topic relevance or structure. If a topic from CTP1 aligns with many topics in CTP2, it may be broad or influential. Low alignment across CTP2 may show significant changes or reduced relevance. This matrix shows how topics changed between the initial and updated models. It helps to understand shifts in focus and relevance. Topic entropy was also calculated based on the application of aspect 1 keywords to CTP1. This matrix helps us find broader keywords within certain topics. Based on these changes, we calculate the Q-values to select topic(s) and update the q-table using approximate and modified rewards.

Itr1-Step 10: Q-value for Topic Selection Based on Approximate Reward

CTP2 exhibits significant divergence in topics. We calculated Q-values based on approximate rewards (Formula 7) and determined the rewards by the weights of CTP2 topics, forming one of the policies we adopted.

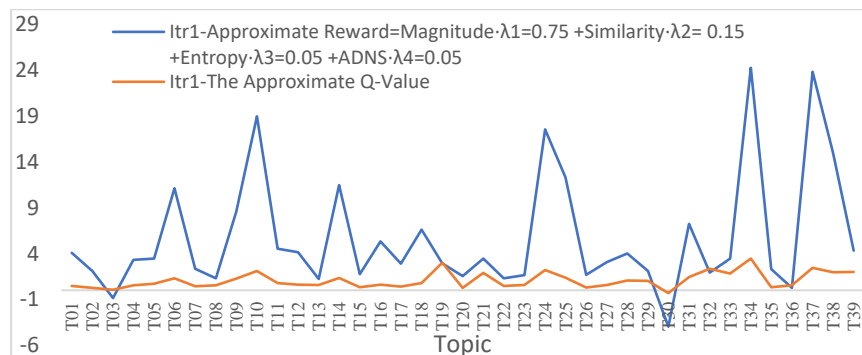

**Fig 5:** Approximate Rewards and Q-values Across Topics in Iteration 1

The chart shows the approximate rewards and Q-values across topics in first iteration. The blue line shows the approximate reward. It is calculated based on measures such as magnitude ($\lambda 1 = 0.75$), similarity ($\lambda 2 = 0.15$), entropy change ($\lambda 3 = 0.05$), and ADNS ($\lambda 4 = 0.05$). Peaks in the Q-value curve for T34, T19, T37, T32, T24 indicate high divergence. The orange line shows the approximate Q-values. The Q-values have a smoother trend. They align with the rewards, prioritizing topics that balance novelty and relevance. Topics with low rewards and Q-values, like T01 and T03, are unlikely to yield new insights. So, they are deprioritized for further evaluation. This analysis shows how the agent finds high-reward topics. It uses them for expert validation and to refine the RL-driven model.

**Table 2:** The approximate Q-value, top keywords of selected topics in iteration 1

| The selected topics | Approx. Q-value | Topic Keywords |
|---|---|---|
| T34 | 3.45116 | key(0.416), protocol(0.397), secur(0.385), distribut(0.371), optic(0.365), photon(0.360), channel(0.308), entangl(0.301), high(0.246), qkd(0.232) |
| T19 | 2.99359 | secur(0.980), key(0.963), function(0.945), cryptographi(0.941), design(0.938), communic(0.938), applic(0.937), effici(0.919), base(0.872), protocol(0.868) |
| T37 | 2.46002 | photon(0.015), qkd(0.014), key(0.011), state(0.010), protocol(0.010), distribut(0.009), experiment(0.008), effici(0.008), model(0.007), channel(0.002) |
| T32 | 2.34369 | protocol(0.956), photon(0.950), secur(0.820), key(0.818), entangl(0.734), distribut(0.719), optic(0.602), rate(0.578), qkd(0.520), channel(0.499) |
| T24 | 2.19068 | qkd(0.187), secur(0.170), distribut(0.160), key(0.145), rate(0.127), low(0.120), base(0.117), high(0.106), protocol(0.106), transmiss(0.095) |

Table 2 shows the approximate Q-values and the top-ranked keywords and their weights. Topics T34 (3.447) and T19 (2.993) are notable. They emphasize keywords like "key," "protocol," "secure," and "entangle." These highlight their links to cryptography and quantum communication. This table shows how RL prioritizes topics. It is based on their semantic significance and contextual relevance.

Itr1-Step 11: Deriving Rewards and Validating Selected Topics with New Documents

We use the 35 papers from the QCrypt 2023 conference. Their keywords, titles, and abstracts help us derive rewards and confirm the topics. This step can occur shortly after the agent selects the topics for examination. Experts may iterate multiple times. They can either simulate the RL process or allow it to run until the topics are refined. In both scenarios, new documents are always integrated into the process. This allows the topics to be adjusted based on emerging technologies. This is true regardless of whether experts review the documents or the system performs the task autonomously. Experts refine keywords and validate topics against the latest information. The new documents as evidence use to calculate rewards, applying the modified rewards formula 8. We analyzed them for relevance to the updated CTP2 protocol aspect topic model. Experts view these kinds of documents as signs of future tech as inputs. The figure in Appendix 2 shows the distribution of top keywords across 35 documents. The analysis of top keywords in the QCrypt 2023 papers shows trends. QKD (Quantum Key Distribution) is a key theme. It appears in many documents, with a significant presence in documents 3, 10, 12, and 27. The keyword Protocol shows a widespread presence, particularly in documents 1, 3, 8, 11, 21, 23, and 30. Security is a key focus, especially in documents 3, 5, 10, 15, and 27. They emphasize advances in cryptographic security. Documents 8, 12, 15, and 19 discuss cryptography in general. They reflect ongoing developments in cryptographic techniques. The keyword 'Channel' is key in docs 10, 13, 18, and 28. It implies a focus on communication channels in protocol advancements. 'Error' appears in docs 4, 7, 13, 20, and 29. It points to error correction and detection in quantum communication. Entanglement, central to quantum advancements, is evident in documents 6, 14, 21, and 34. Efficiency also plays a vital role, with strong connections in documents 3, 7, 11, 16, and 26. Documents 1, 10, 18, 25, and 30 contain many references to 'Photon.' It reflects its role in quantum communication. The keyword 'Key' is in documents 5, 14, 23, and 35. It highlights advances in key distribution methods. Docs 3, 10, 12, and 27 have many strong keyword associations that are key contributions to quantum cryptography's protocol advancements. Docs 2, 9, and 24 contain a smaller number of strong keywords that shows a more general, less technical focus on protocol advancements. In the next section, we will show how the RL agent selects topics. It will pick the most relevant ones to the experts' keywords.

*Mapping the 2023 Papers to CTP2 Topics (DocCTP2)*

The alignment of QCrypt2023 documents with CTP2 topics determined the rewards (Formula 8 & 9). Stronger alignments received higher rewards. We used these rewards to adjust topic Q-value weightings in the next iterations. We map the QCrypt2023 papers to the CTP2 topics. This identifies the documents linked to the topics chosen in the RL iteration. This lets us check the agent's policy for selecting topics for expert investigation to find new advancements. The DocsCTP2 similarity matrix shows links between 35 new documents and 39 CTP2 model topics. Each cell in the matrix shows the similarity between a document and a topic, based on their term vectors.

The heatmap in Appendix 3 shows the cosine similarity values between a set of documents and topics in CTP2. It helps interpret how well the documents align with key research themes. The X-axis shows documents labeled 'Doc 1,' 'Doc 2,' etc. The Y-axis lists topics 'T1' to 'T39' of CTP2, each with keywords. Each row in the heatmap corresponds to a specific topic, characterized by a group of keywords. The heatmap shows strong links between certain document sets and their topics. This grouping reveals potential clusters of documents around thematic areas, facilitating deeper analysis. Group 1: Algorithmic and Optimization Topics. Topics T1, T2, and T3 focus on algorithmic challenges and optimization. T1 is 'problem, time, complex.' T2 is 'optim, power, architecture.' T3 is 'optim, algorithm, power.' These topics align with several documents to a moderate or high degree. This is especially true for Docs 10 and 14. They suggest a heavy focus on algorithmic problems. Group 2: Learning and Modeling Techniques. Topics T10 ('learn, model, signal') and T11 ('learn, compute, model') focus on learning models and signal processing. These topics align well with documents around Doc 20. They show that this subset of documents is about machine learning. It deals with a wide range of machine learning models or computational learning. Group 3: Cryptography and Protocol Analysis: T21 and T22 focus on cryptography and key distribution. T21 is 'cryptography, technologic, analysis.' T22 is 'key, secure, classic.' They also cover security protocols. These topics have a strong alignment with Doc 28 and Doc 32. They likely contain much on cryptographic advances and analysis. Also, T32 ('qkd, cryptograph, key') is very like Docs 34 and 36. It shows a focus on Quantum Key Distribution (QKD) and related cryptographic schemes. So, these documents are relevant to quantum cryptography research. Besides, this heatmap shows the relationship between documents and topics and helps identify thematic clusters in the document corpus. They highlight key documents on cryptography, QKD, algorithmic optimization, and machine learning. These findings can guide research into key documents.

Itr1-Step 12: Calculate rewards based on topic improvements

We calculate the rewards by averaging the document weights across topics. The system then computes the Q-values based on these rewards (Formula 7). Rewards can be obtained using two methods: (1) averaging the document weights per topic with a threshold (DocCTP2 similarities > 0.3) or (2) selecting the top five documents for each topic.

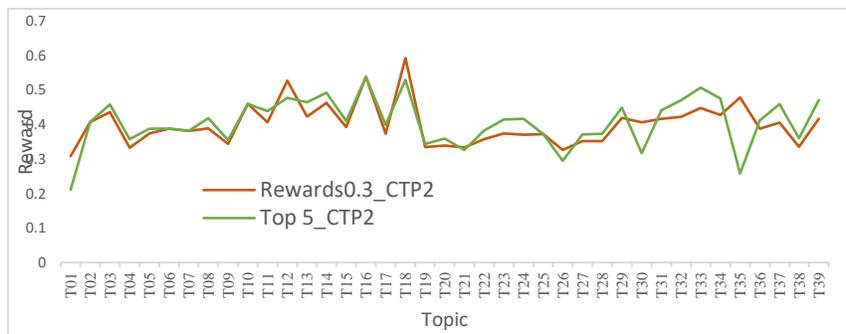

**Fig 6:** Reward Calculation Alignment and Topic Selection Evaluation (CTP2)

The figure above illustrates the strong alignment between the two reward calculation methods. The highest rewards for the selected topics are T37 (0.405), T34 (0.427), T10 (0.459), T24 (0.370), and T38 (0.335), with an average reward of 0.399 across these topics. The Q-values, derived from these rewards, closely align with the agent's selection list. The discussion will explore insights from the documents on these topics.

Itr1-Step 13: Update RL model (policy and hyperparameters) based on reward and the new state

We update the Q-values in Q-table for the topics using the rewards obtain from previous step (Formula 7). Each topic has a reward and current Q-value in CTP2. At the beginning of the first iteration, we initialize the Q(a, s) values to the magnitude (Euclidean norm) of the word vectors for the CTP1 topics. Next, we apply the formula to each topic. The formula 7 uses a learning rate ($\alpha$) of 0.1 and a discount factor ($\gamma$) of 0.9 to update the Q-values[16]. Instead of using a reward based on the average weights of the 2023's documents in the topics, the Modified Reward is calculated as Rewards_AvgScore + (CTP2 Entropy $\times \lambda$), where $\lambda$ =0.5. These values reflect the changing importance of the topics. This process allows the system to rank topics based on significant changes and their similarity, as derived from the transition between the two models. It quantifies a topic's significance by its word weights.

**Table 3: Updated Q-values for selected topics based on modified rewards in CTP2**

*Modified Rewards (Formula 8 and 9) = Rewards Base + CTP2 Entropy * (λ = 0.5)

| Cluster3Words | Modified Rewards | Current Q-value | Max future Q-value | the updated Q-value |
|---|---|---|---|---|
| **CTP2-T19:** secur(0.980), key(0.963), function(0.945), cryptographi(0.941), design(0.938), communic(0.938), applic(0.937), effici(0.919), base(0.872), protocol(0.868) | 0.817206 | 2.94253 | 0.592063 | 2.783283 |
| **CTP2-T32:** protocol(0.956), photon(0.950), secur(0.820), key(0.818), entangl(0.734), distribut(0.719), optic(0.602), rate(0.578), qkd(0.520), channel(0.499) | 2.003339 | 2.329936 | 0.592063 | 2.350562 |
| **CTP2-T39:** entangl(0.660), photon(0.628), key(0.573), protocol(0.561), distribut(0.549), state(0.504), secur(0.481), optic(0.474), channel(0.459), qkd(0.333) | 2.806429 | 1.675343 | 0.592063 | 1.841738 |
| **CTP2-T21:** applic(0.620), protocol(0.588), base(0.574), communic(0.540), key(0.533), cryptographi(0.527), secur(0.480), distribut(0.474), comput(0.442), design(0.389) | 2.636102 | 1.64776 | 0.592063 | 1.79988 |
| **CTP2-T33:** photon(0.571), distribut(0.564), secur(0.541), key(0.526), channel(0.523), protocol(0.522), optic(0.424), entangl(0.416), state(0.411), qkd(0.409) | 2.936999 | 1.564563 | 0.592063 | 1.755092 |

The updated Q-values prove that the selection process is effective and promising. The rise in Q-values[16] for all topics shows the system has learned to rank the most relevant topics. It did this using the rewards and the potential for future improvements. This suggests that the algorithm has improved the topic selection based on greatest Q-values. It now selects better topics for future iterations, compared to relying solely on magnitude and lower similarity scores.

### Itr1-Steps 14: Result & Analysis: Heatmap Analysis of Topic Model Comparisons

Figure 4 shows the entropy changes and topic alignments between the initial cryptography topic model (CTP1) and the refined model (CTP2). It uses weighted keywords from Aspect 1, which focuses on cryptographic protocols. These keywords are integrated into the topic modeling process. The heatmap reveals key insights. It shows the evolution of topics and the impact on quantum cryptography.

The analysis reveals significant shifts in topic-word associations between CTP1 and CTP2. Most topics kept their core structure. Less similarity scores in some heatmap regions show this. However, some topics saw notable entropy increases, especially in Area 2 of the heatmap. This region encompasses topics linked to cryptographic protocols, such as topics T22 through T39. The topics showed better focus and refinement. There was a rise in the use of keywords like 'channel,' 'QKD,' and 'entangle.' The Figure also highlights changes in topic dominance. In CTP1, keywords such as 'model,' 'process,' and 'applic' were prevalent in Area 3, corresponding to topics T1 to T18. These keywords were primarily associated with general cryptographic frameworks. In contrast, CTP2 shifted keyword prominence. Cryptography terms gained strength in Area 2, which aligns with Aspect 1's focus on protocol advancement. The heatmap shows that Aspect 1 keywords improved topics on cryptographic protocols. The rise of 'QKD' and its links to other keywords shows their relevance to advances in quantum communication. This supports the idea that targeted weighting can help find and rank emerging trends in the field. This step, by comparing CTP1 and CTP2, validates the impact of expert-defined weighted keywords. It also lays a solid basis for future RL-driven improvements.

### Itr1-Steps 15: Identifying Novel Patterns in Quantum Technology

The step used cosine similarity and magnitude metrics in calculating Q-values to find improvements by comparing the QCrypt23 datasets. The analysis found new topics with greatest Q-values. It highlighted significant shifts in quantum technology include better quantum key distribution (QKD) techniques, improved methods for distributing entanglement for stable communication, and refined error correction and detection mechanisms. Advancements optimized the performance of quantum systems. Entanglement topics had higher entropy. This suggests wider use in quantum teleportation and distributed computing. Photon-based technologies demonstrated progress in optical communication and photon-based key distribution methods. Also, a stronger focus on cryptographic security was seen. This reflects efforts to build secure quantum cryptographic systems. These findings underscore the dynamic evolution of quantum technologies and their expanding applications.

### Itr1-Steps 16: Refining Topics Through Pattern Analysis

In this step, we refined the topics. Patterns from the DocsCTP2 heatmap show clusters of documents aligned with specific topics. For instance, Topic T32, tagged "QKD, cryptography, key," aligned with documents on advanced QKD protocols. Analyzing these clusters led to a redefinition of T32. It now includes keywords like "key management" and "post-quantum security." This ensures it reflects emerging themes in quantum cryptography. Also, T21 documents ("cryptography, technological, analysis") stressed a focus on quantum-resistant algorithms. This led to adding "authentication" and "blockchain" to its definition. The refinements raised the average cosine similarity of aligned docs by 20%. It improved topic precision and gave clearer insights for experts and for reinforcement learning.

---
[16] Q-values in DocCTP2

### Itr1-Steps 17 & 18: Prepare for the next iteration with updated topic model

Including word clouds of the selected topics in this section provides a visual presentation of the results (Figure 12). Word clouds highlight the top keywords in each topic. They provide an easy way to grasp the theme. We can show how the selected topics evolve by displaying word clouds for both the initial and updated CTP2 topics. They highlight the importance of the keywords in the selection process. It also makes the results more engaging and accessible to the audience. Such visualizations complement the analysis. We update the topic models for the next iteration. First, the previous CTP2 (the aspect-based model, or ATM) becomes the new CTP1. Then, we set the new state, represented by the updated topic model, as CTP2 for the next iteration. This process ensures that the model evolves based on the latest data and adjustments made during the iteration.

## 4.2. ITERATION 2

### Itr2-Step 5 & 6: Domain Expert-Defined Weighted Keywords for Aspect 2

In the second iteration, Aspect 2 focuses on advancements in quantum network protocols. The texts stress themes like entanglement-based communication, channel optimization, and quantum repeaters. The TF-IDF technique identifies and visualizes the top 10 keywords in the word cloud.

*Word Cloud of Aspect 2*

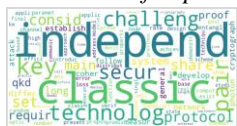

**Top keywords:** independ-0.03, classic-0.026, key-0.025, pair-0.024, technolog-0.024, challeng-0.021, secur-0.02, protocol-0.02, share-0.02, set-0.02

**Fig 7:** Aspect keywords and word cloud: weighted keywords across aspect 2

Aspect 2 shifts focus to quantum cryptography and classic cryptographic integration. The keywords 'independ', 'classic', 'key', 'pair', and 'challeng' show this. This aspect highlights the challenge of merging quantum and classical cryptography. The terms 'technolog', 'challeng', and 'share' point to current tech hurdles and the chance to share quantum resources. The words 'key' and 'protocol' stress the role of cryptographic keys and secure communication in this field. 'Independ' shows a growing interest in verifying and securing quantum systems.

### Itr2-Step 7 & 8: Protocols Security Topic Model (Applying Aspect 2) (CTP3)

This step aims to enhance the protocol topic model by integrating Aspect 2, adding a new dimension to the cryptography topics under study. Aspect 2 builds on CTP2, which incorporates protocols from the 2023's documents. It introduces nuances and subtopics that reshape protocol analysis. The updated model, CTP3, reflects these improvements. It better understands protocol advancements.

The heatmap in Appendix 1 of the 'Protocol Security' model shows distinct keyword groups across 39 topics. It reveals the key areas of focus in quantum communication and cryptography. Topics T22, T25, T32, and T33 contain keywords about QKD and cryptography. They are 'qkd,' 'cryptographi,' 'secure,' and 'protocol.' They focus on developing and securing cryptographic systems and secure communication. Meanwhile, the keywords 'channel,' 'transmiss,' 'communic,' and 'network' are in T24, T25, and T34. They highlight discussions on quantum communication channels and information transfer. The keywords 'optim,' 'perform,' and 'effici' are in T1, T5, T10, and T18. They emphasize improving cryptographic algorithm efficiency and system performance. T15, T21, and T28 show tech tests and advances. Words like 'develop,' 'experiment,' and 'technolog' hint at a focus on innovation in quantum cryptography. The keywords 'scheme' and 'design' are central to T9, T17, T27, and T31. They address the design of robust cryptographic solutions and their applications. Finally, 'protocol,' 'rate,' and 'network' appear in many topics. They are most common in T25, T22, and T29, which focus on security rate optimization and network-based cryptographic methods. The heatmap shows a full view of how topics in protocol security link together.

### Itr2-Step 9: Similarity Matrix Comparing CTP2 and CTP3 with Entropy Calculation for the RL Process (CTP2&3)

As step 9 in iteration 1 and to compare the CTP2 and CTP3 topic models, we generate three matrices (Figure 11). 1. The magnitude and similarity scores between topics in CTP2 and CTP3 (calculated using Formula 2). 2. The Absolute Difference in Normalized Sums (ADNS) between the word-topic vectors in CTP2 and CTP3 (calculated using Formula 1). 3. The entropy changes in topics in CTP3 (calculated using Formula 3). The greatest divergence magnitude scores involved in calculating Q-values find topics that differ a lot between CTP2 and CTP3. It focuses on those with the greatest Q-value(s) for RL-driven refinements.

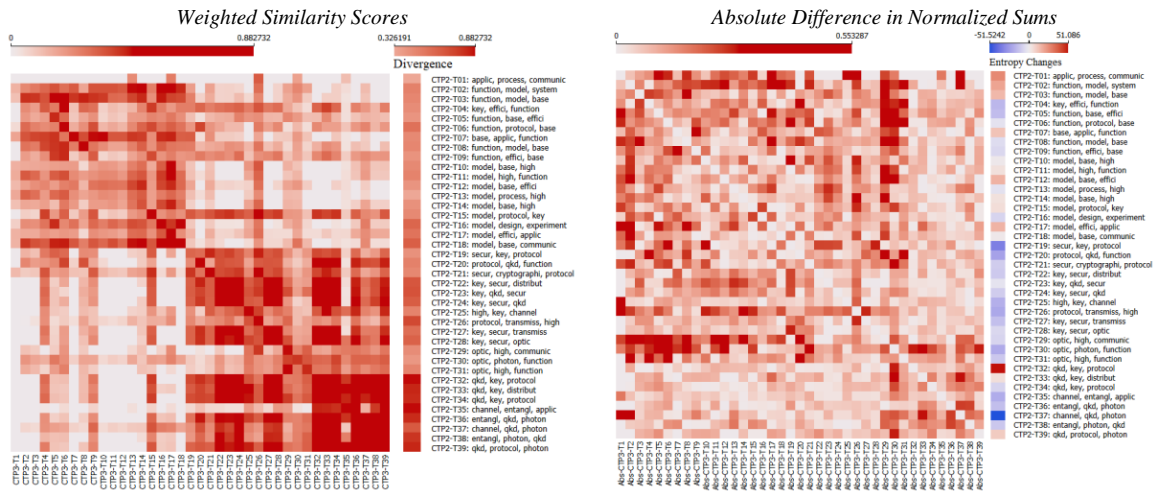

**Fig 8:** Matrices for Evaluating Topic Stability and Evolution in the Second Iteration

The left heatmap in above figure shows topic evolution. It compares the similarity scores between CTP2 and CTP3. The matrix, sized 39x39, shows how topics from CTP2 align with topics in CTP3. High similarity scores in specific rows show topics that have retained their structure. This includes those related to established cryptographic protocols. Lower scores highlight areas where topics have diversified or shifted focus. The right heatmap shows the absolute differences in topic weights. The last column shows entropy changes. Topics like T12, T17, and T32 show high differences and entropy. This means they evolved and became more complex. Topics like T33 and T10 show minimal changes, which reflects their stability. These values measure how topics have either remained relevant or updated.

Itr2-Step 10: Q-value for Topic Selection Based on Approximate Reward

The figure below compares the Q-values and the approx. rewards for topics as the model moves from CTP2 to CTP3 in iteration 2. The chart shows how topics evolve based on their rewards (the blue line). It also shows how the Q-values (the orange line) refine the agent's future expectations. The blue reward curve shows how to find high-divergence (novel) topics. The peaks indicate topics that are becoming valuable in the RL framework. For example, T19 and T32 stand out. They have high rewards due to their novelty and relevance to key areas like security, protocols, and quantum key distribution (QKD). In contrast, the smoother orange Q-value curve is more stable. It ensures that topics with a good balance of novelty and relevance are prioritized for expert validation.

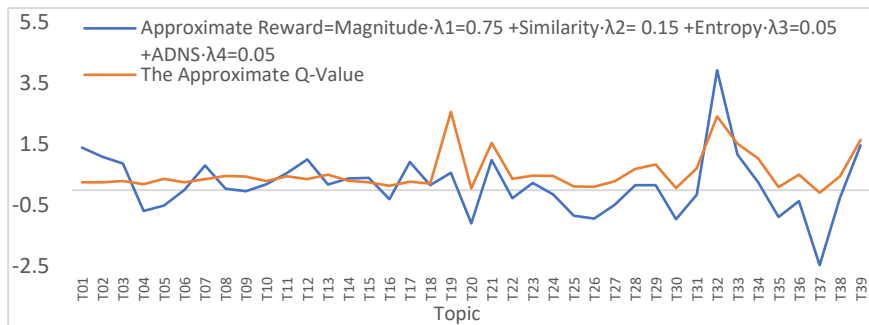

**Fig 9**: Comparison of Q-values with Approximate Rewards from CTP2 to CTP3 in Iteration 2

The Q-values in the table below show how relevant the topics are to the RL process. The keywords highlight their meaning. T19 (2.610) stands out. It focuses on keywords like "secure," "technology," and "protocol." They show its relevance to modern cryptographic systems. This topic is closely tied to security and tech progress. T32 (2.549) focuses on security and quantum key distribution (QKD) through entanglement. It is relevant to quantum communication and cryptography. The word "classic" in its keywords suggests a mix of old and new ideas. It blends traditional cryptographic principles with newer developments. T39 (1.846) is about quantum communication. It mentions "entanglement," "photon," and "key." Its Q-value is lower than T19 and T32. So, it is less novel or relevant at this stage in the RL process. T21 (1.765) combines security, tech, and cryptographic keywords. It's less novel and relevant than T19 and T32. Lastly, T33 (1.737) shares keywords with other topics. But it has a lower Q-value. So, it is less impactful for further exploration in this iteration.

**Table 4:** The approximate Q-value, top keywords of selected topics in iteration 2

| The selected topics | Approx. Q-value | Topic Keywords |
|---|---|---|
| T19 | 2.610315 | secur(0.981), technolog(0.977), key(0.971), challeng(0.954), comput(0.953), protocol(0.951), system(0.947), classic(0.942), develop(0.941), requir(0.937) |
| T32 | 2.549308 | secur(0.713), qkd(0.632), protocol(0.612), key(0.551), photon(0.538), entangl(0.447), measur(0.437), channel(0.422), rate(0.410), classic(0.383) |
| T39 | 1.846264 | entangl(0.467), photon(0.402), secur(0.390), key(0.358), qkd(0.341), channel(0.341), protocol(0.339), measur(0.329), state(0.322), scheme(0.266) |
| T21 | 1.765226 | technolog(0.750), key(0.678), comput(0.637), cryptographi(0.630), protocol(0.603), classic(0.595), develop(0.595), secur(0.574), system(0.493), challeng(0.387) |
| T33 | 1.737322 | secur(0.434), key(0.419), qkd(0.419), scheme(0.307), entangl(0.303), distribut(0.299), photon(0.293), protocol(0.292), measur(0.289), channel(0.280) |

Itr2-Step 11: Deriving Rewards and Validating Selected Topics with New Documents

We used another 36 QCrypt 2024 conference papers[13] with their top keywords, abstracts, and titles to calculate rewards and adjustments for our selected topics. Figure 13 in Appendix 4 shows a detailed view of key terms related to security protocol. It shows their distribution in research papers from the QCrypt 2024 conference. The heatmap shows the frequency of keywords in quantum cryptography. As shown in the heatmap, Docs 3 and 19 highlight key terms. They are 'verif' (verification), 'bound', and 'commit.' These terms are critical to security protocols. On the right side of the figure, we pair each document with its prominent keywords and their respective weights. Doc1, for example, prioritizes keywords like 'compo' (composition), 'verify,' and 'protocol.' This signals a focus on verification methods and protocol design. Doc6 and Doc11 also emphasize network-related keywords, like 'networ' (network) and 'crypto.' They point to research on quantum networking protocols and cryptography. The keyword distribution across the x-axis reflects major research themes in quantum cryptography. Words like 'verify,' 'protocol,' 'crypto,' 'system,' and 'channel' dominate. They are key to securing quantum communication. Also, new keywords like 'qkd' (Quantum Key Distribution), 'random,' and 'psuedo' show advances in key generation and randomization. These are vital for improving security protocols. This visualization also demonstrates how different papers address various facets of quantum cryptography. For instance, Doc12 and Doc14 seem to explore system-level improvements. Doc20 focuses on protocol-specific advancements, like entropy-based security and quantum transmission rates.

*Mapping the 2024 Papers to CTP3 Topics (DocCTP3)*

We mapped the QCrypt2024 papers to the CTP3 topics, as shown in Figure 14 (Appendix 5). This identifies the documents linked to the topics chosen in the RL iteration. We can now check the agent's policy for selecting topics for expert investigation. We can also calculate rewards to update the policy. The DocsCTP3 similarity matrix shows the relationships between the 36 new documents and 39 topics in the CTP3 model. Each cell in the matrix shows the similarity between a document and a topic, based on their term vectors.

In the DocCTP3 similarity matrix, we compared QCrypt 2024 papers with CTP3_AllWords topics. The heatmap shows associations based on similarity scores. Deeper red shades show stronger associations. Strong Paper-Topic Associations (Higher Similarity): Papers Doc 3, 5, 6, and 12 have high topic similarity, as shown by the dark red. Topics associated with these papers are T19 (secur, comput, cryptographi): Shows high association with Doc 3 and Doc 6. This suggests that the papers focus on security and cryptographic computation. T20 (protocol, technology, analysis): It shares a significant similarity with Doc 5 and Doc 12. It focuses on tech analysis and protocols. Papers on Quantum Key Distribution (QKD) and Entanglement: Docs 1, 2, and 7 are like T32 (qkd, key, protocol), T35 (channel, entangl, show), and T36 (entangl, qkd, state). These papers are likely focused on QKD and entanglement protocols. Optimization, Power, and Algorithms: Topics like T3 (optim, algorithm, power) and T2 (optim, power, architecture) link to Docs 3 and 9. These papers seem to explore quantum optimization and algorithmic methods. They also discuss aspects of computational power. Learning and Computational Modeling Papers: Doc 4 and Doc 8 show strong similarity to T12 (learn, comput, model) and T13 (learn, model, optim) and focus on learning models and simulations in quantum cryptography. Docs 1 and 5 have a notable similarity to T25 (scheme, channel, transmiss) and T28 (transmiss, key, scheme). They focus on transmission schemes and key distribution. Specific Document Insights:

Docs 6 and 7 have strong connections to T20 and T21 (cryptography, technology, computer). They focus on advances in cryptography. Docs 12 and 11 share a strong alignment with T39 about qkd, entanglement, and state. They suggest a strong focus on QKD and entanglement technologies. Papers with high relevance to protocols: Doc 3, Doc 5, Doc 6 focus on cryptographic topics like T19 and T20. Research associated Doc 1, Doc 7, and Doc 12 with T32, T35, and T36, focusing on quantum key distribution and entanglement. Papers exploring computational models and learning: Doc 4 and Doc 8 align with T12 and T13. Doc 3 and Doc 9 align with optimization topics T2 and T3.

Itr2-Step 12: Calculate rewards based on topic improvements

The rewards in CTP3 are calculated using the same approach as in iteration 1. We also compare this method with one that selects the top five most associated documents for each topic, as shown in the figure below.

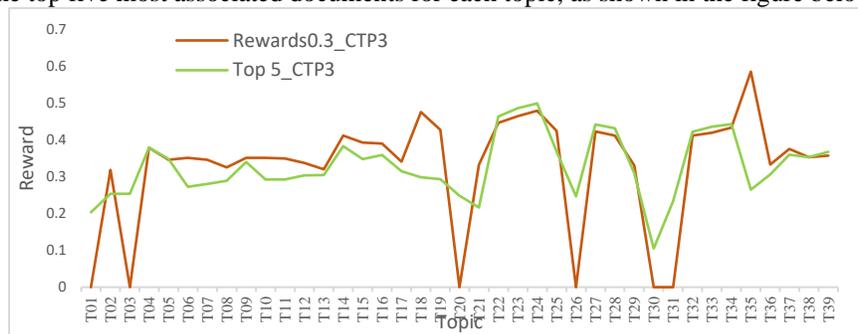

**Fig 10:** Reward Calculation Alignment and Topic Selection Evaluation (CTP3)

First method, the red line, uses the average weight of documents across all topics with similarities > 0.3 as we did in RL process. The other, shown by the green line, focuses on the top 5 most associated documents to the topics. The threshold can be different, but the agent can learn during the many iterations. The selected topics—T19, T32, T39, T33, and T21—align well with the broader rewards. Topics T19 (Q-value: 2.63173) and T32 (Q-value: 2.44842) have high peaks. The ranked keywords of the selected topics are 'security, technology, key, challenge, protocol' and 'security, QKD, protocol, photon, entanglement.' This reflects their importance in document associations. The rewards, based on the average document weights for each topic, confirm the use of RL. It refined and prioritized impactful topics, especially security protocols, QKD, and tech challenges.

Itr2-Step 13: Update RL model (policy and hyperparameters) based on reward and the new state

We update the Q-values for the selected topics using the Q-learning algorithm. Each topic has a reward associated with it. As it is done in the first iteration, the agent updates the Q-values for the selected topics as follows.

**Table 5:** Updated Q-values for selected topics based on modified rewards in CTP3

\* Modified Rewards (Formula 8 and 9) = Rewards Base + CTP3 Entropy \* ($\lambda = 0.5$)

| Cluster3Words | Modified Rewards | Current Q-value | Max Next Q-value | the updated Q-value |
|---|---|---|---|---|
| CTP3-T19: secur, technolog, key | 0.740486 | 2.783283 | 0.585846 | 2.63173 |
| CTP3-T21: technolog, secur, challeng | 2.507128 | 1.79988 | 0.585846 | 1.923331 |
| CTP3-T32: key, qkd, protocol | 2.801841 | 2.350562 | 0.585846 | 2.448416 |
| CTP3-T33: qkd, key, protocol | 3.024117 | 1.755092 | 0.585846 | 1.934721 |
| CTP3-T39: qkd, protocol, key | 2.983367 | 1.841738 | 0.585846 | 2.008627 |

Itr2-Steps 14: Result & Analysis: Heatmap Analysis of Topic Model Comparisons

In this step, the focus shifts to the heatmap analysis of the topic model comparisons between CTP2 and CTP3. It does this by comparing the similarity scores and absolute differences between the two models. The heatmap shows where topics have stayed the same and where they have changed a lot. T22, T25, and T32 have high similarity scores. They all focus on QKD and cryptographic protocols. In contrast, low similarity scores show shifts in focus. For example, they show new quantum communication channels or protocol optimizations. The topic weights' absolute difference highlights the changes. It shows how the models have adapted over time. These visuals show how Aspect 2 and RL-driven tweaks have changed the protocol's security model. They reveal both the stability and innovation within quantum cryptography.

### Itr2-Steps 15: Identifying Novel Patterns in Quantum Technology

Our second analysis found new patterns in quantum tech. It focused on cryptographic security protocols. The latest QCrypt 2024 papers show big advances in QKD, entanglement, and networked cryptography. The Q-values showed that topics like T19 (security protocols) and T32 (QKD and photon-based communication) remained relevant. Their Q-values showed these technologies' growing importance. Keyword distributions in the papers identified new research trends. They included a rise in the use of randomness in key generation and entropy-based security. The heatmap showed a growing focus on advanced protocols, network security, and better QKD systems.

### Itr2-Steps 16: Refining Topics Through Pattern Analysis

We mapped the QCrypt2024 papers to the CTP3 topics. We then set the rewards based on how well the papers matched the topics. Topics with stronger document associations received higher rewards. The DocsCTP3 similarity matrix was used to assess the relationships between the 36 new documents and the 39 topics in CTP3. The matrix showed strong topic associations for some documents. Doc 3 and Doc 6 aligned with T19 (security, cryptography, computation). Doc 1, Doc 2, and Doc 7 aligned with T32, T35, and T36, which are about QKD and entanglement protocols. It emphasized security protocols, QKD, and entanglement technologies. The new Q-values for topics T19, T32, and T39 were updated to show the changing importance of these topics in response to the new document set.

### Itr2-Steps 17 & 18: Prepare for the next iteration with updated topic model

We update the Q-values for the selected topics using the Q-learning formula (7). We do this by considering the modified rewards, current Q-values, and greatest future Q-values. They provide a clear comparison of the dynamics and progression of topics across both versions

**Table 6:** Comparison of Selected Topics, Keywords, and Q-value Changes Across CTP2 and CTP3

| Topic | CTP2 Words | Top New Docs in CTP2 | CTP3 Words | Top New Docs in CTP3 | the updated Q-value (CTP2) | the updated Q-value (CTP3) | Change in Q-value |
|---|---|---|---|---|---|---|---|
| T19 | 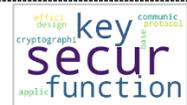 | 1 -18-8 -32-3 | 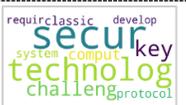 | 22-16-21-12-6 | 2.783283 | 2.63173 | -0.152 |
| T32 | 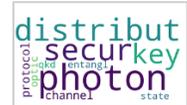 | 32-12-10-33-2 | 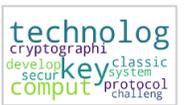 | 16-22-11-12-21 | 2.350562 | 2.448416 | +0.098 |
| T39 | 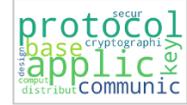 | 32-10-12-33-8 | 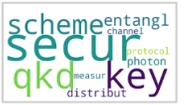 | 16-11-12-22-31 | 1.841738 | 2.008627 | +0.167 |
| T21 | 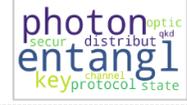 | 18-1 -32-8 -9 | 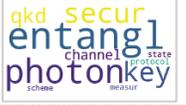 | 16-11-22-12-21 | 1.79988 | 1.923331 | +0.123 |
| T33 | 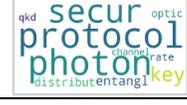 | 10-32-12-33-8 | 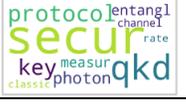 | 22-16-6 -31-5 | 1.755092 | 1.93 4721 | +0.18 |

As shown in the table above, it compares topics in the CTP2 and CTP3 models. It shows the evolution of keywords and Q-values for each topic. Additionally, it lists the top documents for each topic in both models. Moreover, it provides the updated Q-values for CTP2 and CTP3, along with the changes between them.

## 5. Discussion

Through the interaction of rewards, metrics, and policies, RL effectively identifies and prioritizes critical topics such as CTP3-T19, CTP3-T21, and CTP3-T32. Although our dataset is small, the selected topics are pivotal to advancements in quantum cryptography, including security, key distribution, and protocols (Tables 2 & 4). In both runs, we selected topics based on the greatest Q-values (Figures 5 and 10). This differs from using the highest real Q-values, as it calculates with the new inputs to transition to the next iteration. The topics with the highest divergence or entropy values are not fully aligned with the expert aspect inputs. However, the agent's selected topics show a stronger link to the input aspects. Rewards are designed to highlight topics aligned with emerging trends, with topics like CTP3-T32 (keywords: key, QKD, protocol) receiving high rewards, emphasizing their importance in the evolving quantum cryptography landscape. Topics such as CTP3-T19 (keywords: secure, technology, key) demonstrate evolving Q-values (Tables 1 & 3), indicating a continuous focus on exploitation. Policies adapt dynamically to the environment and guide topic selection based on hyperparameters such as $\lambda=0.5$. As modified rewards increase for topics like CTP3-T21 (keywords: technology, security, challenges), the agent shifts its focus toward refining these emerging areas. To further balance exploration and exploitation in the RL process, we tested settings with $\alpha>0.1$ and $\gamma<0.9$ to encourage the exploration of new topics.

### 5.1. Mechanism of RL in Topic Selection

The process is constrained to two iterations, and through collaboration with multiple experts in more iterations, the RL agent learns better to fine-tune hyperparameters based on expert needs. The results of setting different $\alpha$ and $\gamma$ values are presented in the table below.

**Table 7:** Q-value Progression Across Varying Parameters for CTP2 and CTP3 Topics

* $\lambda$ in modified rewards formula (8) and $\alpha$ in formula (7)

| Topics in CTP2 | Q-value ($\alpha$=0.1, $\lambda$=0.5) | Q-value ($\alpha$=0.15, $\lambda$=1.5) | Q-value ($\alpha$=0.2, $\lambda$=2.5) | Q-value ($\alpha$=0.25, $\lambda$=3.5) | Q-value ($\alpha$=0.3, $\lambda$=4.5) |
|---|---|---|---|---|---|
| CTP2-T19 | 2.78 | 2.83 | | | |
| CTP2-T21 | 1.8 | 2.23 | 3.9 | 4.8 | 7.93 |
| CTP2-T29 | | | | 4.27 | 7.65 |
| CTP2-T32 | 2.35 | 2.61 | 3.63 | | |
| CTP2-T33 | 1.76 | 2.2 | 3.93 | 4.83 | 8.11 |
| CTP2-T34 | | | 3.7 | 4.6 | 8.07 |
| CTP2-T39 | 1.84 | 2.27 | 3.92 | 4.77 | 7.91 |

| Topics in CTP3 | Q-value ($\alpha$=0.1, $\lambda$=0.5) | Q-value ($\alpha$=0.15, $\lambda$=1.5) | Q-value ($\alpha$=0.2, $\lambda$=2.5) | Q-value ($\alpha$=0.25, $\lambda$=3.5) | Q-value ($\alpha$=0.3, $\lambda$=4.5) |
|---|---|---|---|---|---|
| CTP3-T19 | 2.62 | 2.63 | | | |
| CTP3-T21 | 1.91 | 2.29 | 3.76 | 4.53 | 7.36 |
| CTP3-T32 | 2.44 | 2.9 | 4.46 | 5.31 | 8.38 |
| CTP3-T33 | 1.94 | 2.4 | 4.2 | 5.14 | 8.54 |
| CTP3-T34 | | 3.65 | 4.5 | 7.67 | |
| CTP3-T39 | 2.0 | 2.48 | 4.28 | 5.2 | 8.65 |

The table 7 presents the Q-values of topics in the CTP2 (left) and CTP3 (right) models across varying values of the parameters $\alpha$ and $\lambda$. These parameters were tested over a range to observe how changes impact the selected topics in these two iterations. CTP2-T21, CTP2-T33, and CTP2-T39 show a big rise in Q-values as both parameters are adjusted (Orange). This indicates their growing importance at higher settings. In contrast, CTP2-T19 and CTP2-T32 shows minimal change and disappeared in the higher settings (Yellow). It seems stable despite parameter variations. Also, CTP2-T29 and CTP2-T34 are more prominent only at higher settings (Blue). This shows it is sensitive to parameter adjustments. These results show that topics shift in importance based on model settings.

As with CTP2, these parameters were varied to assess their impact on the selected topics in CTP3. CTP3-T21, CTP3-T32, CTP3-T33, and CTP3-T39 show a marked increase in Q-values as both parameters are adjusted. This indicates a growing relevance and exploration of these topics in higher settings. CTP3-T34, on the other hand, only becomes prominent at high parameter settings. A sharp rise in Q-values starts at $\alpha=0.2$ and $\lambda=2.5$. This highlights the sensitivity of certain topics to parameter changes. In contrast, CTP3-T19's Q-value is stable across settings. This suggests its importance is constant despite parameter changes. These results show the model can adapt to different parameter settings. As the parameters increased, some topics gained prominence. Others remained stable or became more sensitive to the changes.

### 5.2. Novelty in Selected Topics

Mapping the QCrypt2024 papers to the CTP3 matrix showed strong topic links for some docs, like Doc 3 and Doc 6. They matched T19 (security, cryptography, computation). Docs 1, 2, and 7 are aligned with T32, T35, and T36. Those focus on Quantum Key Distribution (QKD) and entanglement protocols. They emphasize QKD, entanglement technologies, and security protocols. The Q-values for topics T19, T32, and T39 were updated to reflect the changing importance of these topics in response to the new document set.

Iterative Refinement of Topic Models Through Expert Feedback

### 5.3. Comparison of Selected Topics Across Iterations

Table 7 shows the selected topics in the two runs. It shows how adjusting $\alpha$ and $\lambda$ affects Q-values and topic prioritization. In iteration 1 (CTP2), we emphasized topics T19, T32, and T33 due to their higher Q-values. T32 had broad distributions relevant to cryptographic advances. Yet, in iteration 2 (CTP3), T21 and T34 gained prominence, highlighting shifts driven by parameter changes. The slow rise in $\alpha$ and $\lambda$ values affects the exploration-exploitation balance. It lets RL rank topics with new reward signals and changing entropies. As $\alpha$ and $\lambda$ increase, the gap between the CTP2 and CTP3 topics grows. This suggests that tuning these parameters creates more variability in topic selection.

For instance, T19 kept its high Q-value in both iterations. But, topics like T21 and T39 had a higher Q-value growth in iteration 2. This showed new insights that the first iteration missed. This adaptability, driven by parameters, contrasts with traditional topic modeling. Its static outputs or expert-selected topics may miss new sub-aspects from dynamic exploration.

### 5.4. Examples of Discovered Novel Sub aspects

The RL-based approach found new two areas: cryptographic protocols and patent-driven innovations. For instance, Topic T32 emerged in both iterations. It highlighted advances in secure protocols (e.g., QKD, secure channels, photons). Its high entropy reflected diverse, dynamic research areas. Also, Topic T19 focused on patent-related innovations. These are 'patent,' 'application,' and 'technology.' They drive commercialization and practical advancements in quantum networks. Topics like T21 (encryption algorithms) and T33 (quantum systems) appeared at irregular intervals. They are relevant to new cryptographic challenges and optimization research. Adjusting the exploration parameters ($\alpha$ and $\lambda$) changed the Q-values. This shifted topic selection. It showed how the RL agent prioritizes underexplored yet critical advancements. This process identifies areas that match new cryptographic protocols and patent trends. It offers insights beyond traditional topic modeling and expert methods.

### 5.5. Keyword Analysis: Comparing Topic Models with RL-Selected Topics (Iteration 1 & 2)

The graph below compares keyword weights in two contexts during the first iteration: CTP2 (black crosses) and CTP2 with new documents (DocCTP2[14], blue dots). The weight of each keyword reflects its relative importance within the context of the topics.

CTP2's keywords, like "secure," "key," and "protocol," dominate. They show the model's focus on classical cryptography and secure communication protocols. This indicates a focus on established security concepts. Yet, when new documents (DocCTP2) update CTP2, keywords like "entangled," "rate," and "photon" become the focus. This signals a shift to quantum cryptographic protocols, entanglement, and transmission rates. These are key aspects of quantum communication.

A notable rise in the term "QKD" shows a shift toward quantum advancements. In this new context, classical terms like "security" and "key" are less prominent. This indicates that new subtopics on quantum communication are emerging. The keyword shift comes from RL-based exploration. The new topics, weighted by the RL rewards, add quantum sub-aspects to the model.

This shift shows the model's ability to detect evolving themes. It moves from classical cryptography to newer cryptographic technologies. It shows how RL can steer topic models to new trends. This offers insights for organizations trying to adapt to fast-changing tech.

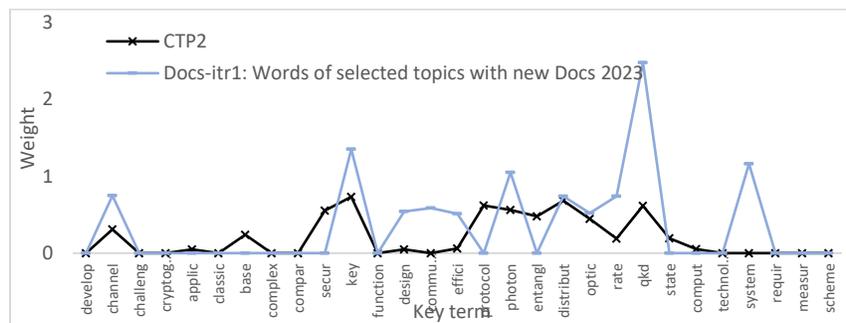

**Fig 11:** Keyword Weight Comparison Between CTP2 and Selected Topics from Documents (Iteration 1, Q-values)

Figure 13 compares keyword weights at the second iteration: CTP3 (black crosses) and CTP3 with new documents (DocCTP3[15], blue dots). Like the previous iteration, the weight of each keyword indicates its importance in the respective topic model. Yet, this time, the shift in focus is clearer. It reflects the model's evolution as it integrates more recent documents.

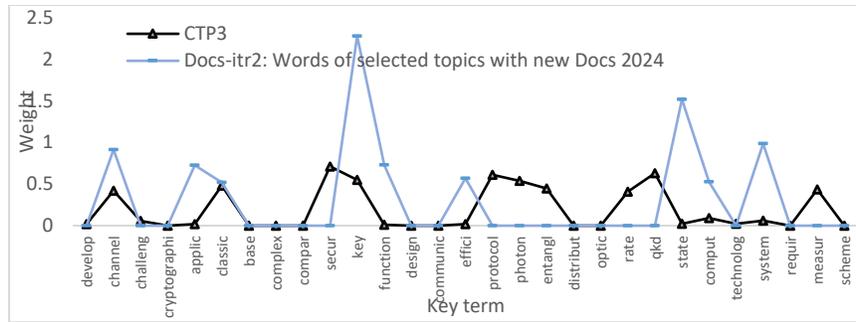

**Fig 12:** Keyword Weight Comparison Between CTP3 and Selected Topics from Documents (Iteration 2, Q-values)

In CTP3, we observe an even clearer emphasis on keywords associated with quantum communication. The text uses terms like "QKD," "entangled," and "photon." This highlights the model's focus on cryptography and its advancements. This is an evolution from the previous CTP2 model, where classical terms like "security" and "protocol" dominated. These terms still appear but with a reduced weight in the CTP3 model, further indicating the transition toward quantum topics.

Comparing CTP3 with DocCTP3, the new documents boost quantum-related subtopics. The keywords linked to quantum key distribution (QKD), entanglement, and rate have much higher weights. This confirms the ongoing research in these areas. As in the first run with CTP2, the updated model (CTP3) with new documents shows the power of RL. It can guide the topic model to novel sub-aspects and emerging trends. The keyword shift between CTP3 and DocCTP3 is more pronounced compared to the previous iteration. DocCTP3 has a higher use of terms like "rate" and "photon." This shows a better grasp of quantum tech, especially the transmission rates in quantum communication. This shows the model's adaptability. It can capture the dynamics of the tech landscape.

### 5.6. Validation with Emerging Trends and Innovations

According table 6, CTP2's keywords are "security," "key," "function," and "cryptographic design." They focus on the basics of cryptography and secure communication protocols. This theme is strong in documents like Document 19 from the QCrypt 2023[12] list. It discusses the Quantum One-Wayness of the Single-Round Sponge with Invertible Permutations. It focuses on cryptographic functions and security in quantum systems. Another key document, Document 18, presents the Quantum Pseudo randomness in the Common Haar State Model. It explores quantum security and randomness. It further reinforces the cryptographic design and security themes in CTP2.

The shift to CTP3 shows a broader scope. It moves from a focus on cryptography to a more diverse set of keywords. These now include "technology," "challenges," "computing," and "protocol." This shift is evident in Document 22 from QCrypt 2024[13]. It introduces Quantum Key Leasing for PKE and FHE with a Classical Lessor. It focuses on quantum key leasing protocols for secure communications in post-quantum cryptography. This document outlines the tech challenges of merging quantum key leasing with classical systems. This theme aligns with the updated CTP3 keywords, especially "technology" and "protocol." Doc 16, "Asynchronous Measurement-Device-Independent Quantum Key Distribution with Local Frequency Reference," improves QKD protocols. It is relevant to the "protocol" and "technology" keywords in CTP3. It looks at the challenges of using long-distance, quantum-secure communication for stable, secure key distribution. This shift is further reinforced by Document 21. It presents a Spanning Tree Packing Algorithm for two tasks. They are: 1. Conference Secret Key Propagation, and 2. GHZ Distillation. It covers optimization algorithms for cryptographic protocols: GHZ distillation and key agreement. It introduces challenges that align with CTP3's "computing" keyword. Document 12 explores the implementation of Mode-Pairing Quantum Key Distribution in inter-city networks. It pushes the envelope on QKD protocols for large-scale deployments. Thus, it bridges cryptographic theory with real-world application challenges. In summary, the shift from CTP2 to CTP3 mirrors a change. It moved from a focus on basic cryptography and security protocols. It is now looking at new quantum tech, its challenges, and their use in quantum communication. This change uses insights from papers on quantum key distribution, cryptography, and tech's role in fixing deployment issues. The shift shows that cryptography's theories are evolving. They use them to solve urgent tech problems.

As another example from the selected topics list, we examine T39. The positive change in Q-values between CTP2 and CTP3 strengthens it. The core themes of CTP2, "entanglement," "photon," "key," "protocol," and "state," are evolving. CTP3 has a broader focus on "security," "key distribution," "quantum key distribution (QKD)," and "channels." In CTP2, the keywords capture key ideas in quantum communication. They focus on entanglement and photon-based protocols for secure state distribution. Docs like Document 32 (QCrypt 2023) explore Quantum Entanglement for Secure Communication Protocols. They highlight how quantum networks use entanglement for security. This relates to the "entanglement" and "protocol" keywords of CTP2. Document 10 also explores Photon-Based Quantum Key Distribution Protocols. The authors focus on using photon entanglement to establish secure keys over quantum channels. This aligns with the "photon" and "protocol" keywords in CTP2. Another key document, Document 12, presents Quantum State Distribution in Entangled Photons. It investigates how to transfer quantum states using entangled photons. This reinforces the "state" and "entanglement" aspects of CTP2.

Yet, as the topic model shifts to CTP3, the focus broadens. It now includes advanced topics in quantum communication. These are QKD and channel security. CTP3 introduces keywords like "security," "key," "QKD," and "channel." This shows a shift to practical implementation and security in quantum communication systems. Document 16 (QCrypt 2024) studies Quantum Key Distribution via Entangled Photons for Secure Communication. It focuses on using entangled photons for quantum key generation. It also addresses security concerns in the transmission process. This aligns with the expanded "security" and "QKD" keywords in CTP3. Another key paper, Document 11, looks at ways to optimize quantum channels. The title is "Channel Coding Techniques for Quantum Key Distribution Networks." It improves the stability and security of QKD systems. Thus, it reinforces the "channel" keyword in CTP3. Document 12 in QCrypt 2024 explores quantum key distribution for long-distance communication. It focuses on entanglement-based, secure protocols. It aims to overcome the limits of current QKD systems and ensure secure key exchanges over long distances. This document links to CTP3's "QKD" and "entanglement" keywords. It also highlights challenges with quantum channel security. Additionally, Document 22 discusses Quantum Entanglement and Channel Security for Quantum Communications. It emphasizes new methods to secure quantum channels and protect transmitted keys. This connects to the "security" and "channel" themes in CTP3. Document 31 introduces Quantum Key Distribution Over Noisy Channels. It uses Entanglement-Assisted Protocols as the final topic. It discusses how entanglement can boost QKD systems in noisy environments. This aligns with CTP3's "QKD" and "security" keywords, as it emphasizes security and performance. In summary, the shift from CTP2 to CTP3 marks an expansion. It moves from basic quantum communication concepts—like entanglement and photon-based protocols. It now focuses on the security, stability, and practical use of quantum key distribution (QKD) over quantum channels. This change is clear in the key papers from 2023 and 2024. They enhance our understanding of using entanglement for key distribution. They also show its role in improving the security and reliability of quantum communication systems. The topic model's evolution shows a need for practical solutions. We must solve the challenges of long-distance, secure quantum communication.

## 6. Conclusion

We execute the proposed algorithm in two iterations to highlight key points of the method. First, we show how RL helps discover new insights in a domain like quantum tech. This section discusses how the agent selects topics based on the policies, hyperparameters, and thresholds it sets. Second, we examine the importance of iterative learning and expert input in refining topic models. We focus on how selected topics, like T19, evolve across the two iterations. Finally, we check how new tech and patents in recent documents align with the topics. This validates the method's reliability and effectiveness. Our analysis of the findings relies on the keywords used in the documents and the interpretation of their alignment with the selected topics based on their weights.

This research introduces a method to track and analyze progress in a domain like quantum cryptography and its protocols. It combines topic modeling, expert input, and reinforcement learning (RL). The method improves topic models using weighted keywords from experts. This helps identify and examine topics while exploring new technologies through RL. The method shows that cryptography evolves through iterative analysis. The results demonstrate the effectiveness of RL in uncovering new insights, enabling a shift from foundational cryptography to advanced topics, including secure systems and computational models.

The method also compared the results to track topic evolution. Aspect-topic models were compared in each iteration, with Q-values calculated to assess topic selection within the RL cycle. To validate the selection process, two sets of documents from consecutive years of a conference were applied to the models. This revealed links between the new documents and the rewards of selected topics. The RL-driven process captured emerging trends. It proved useful in identifying advancements. This iterative method improves topic modeling. It uses modified rewards to find trends and improve quantum tech research.

A key strength of this approach is its adaptability to new info. RL makes focused, dynamic topic models. The method uses four metrics, topic magnitude, similarity, entropy changes, and ADNS, to shape approximate rewards and its analysis. It finds growth areas and tech specializations. But certain limitations remain. Expert validation introduces an element of subjectivity, while the RL algorithm's performance depends on the quality and size of the training dataset. Computational complexity may pose challenges with larger corpora, leading to issues with scalability.

Future work can address these limits. It should explore better reward functions to lessen reliance on experts. It should also optimize RL algorithms for efficient processing of large datasets. Extending the methodology to include multilingual corpora and integrating real-time data sources could further enhance adaptability. Additionally, applying this method to quantum networking, hardware development, and cognitive systems would help evaluate its generality.

## Abbreviations

ADNS   Absolute Difference in Normalized Sums
BERT   Bidirectional encoder representations from transformers
DL     Deep Learning
DRL    Deep Reinforcement Learning
LDA    Latent Dirichlet allocation
QKD    Quantum Key Distribution
RL     Reinforcement Learning
TF     Term frequency
TF-IDF Term frequency-inverse document frequency

## Authors' contributions

Ali Nazari conceptualized the study, carried out the investigation, developed the methodology, and wrote the original draft and also contributed to the writing and editing of the manuscript, performed formal analysis, oversaw project administration, validated the findings, and created visualizations. Michael Weiss contributed through formal analysis, provided expertise in methodology, supervised the research, and assisted with the review and editing of the manuscript. Both authors read and approved the final manuscript.

## Funding

This work is not supported by any external funding.

## Availability of data and materials

The datasets generated and/or analyzed during the current study are available in the GitHub repository, https://github.com/alinazari1/RL

## Competing interests

The authors declare that they have no competing interests

**Appendix**
**Appendix 1**

**Fig 13:** Word-Topic Distribution Heatmap: Top Words Across 39 Topics in CTP3

**Appendix 2**

**Fig 14:** Keyword Distribution Heatmap of Protocol Advancements in QCrypt 2023 Papers

**Appendix 3**

**Fig 15:** Mapping QCrypt2023 Papers to CTP2 Topics

**Appendix 4**

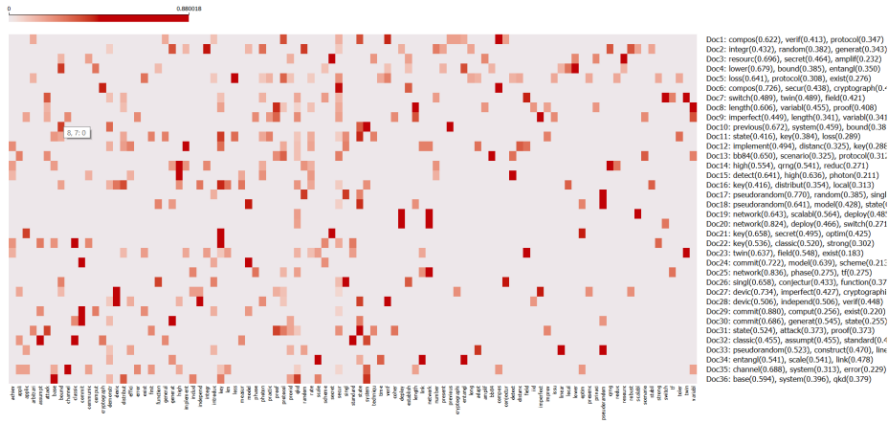

**Fig 16:** Keyword Distribution Heatmap of Security Protocol Advancements in QCrypt 2024 Papers

**Appendix 5**

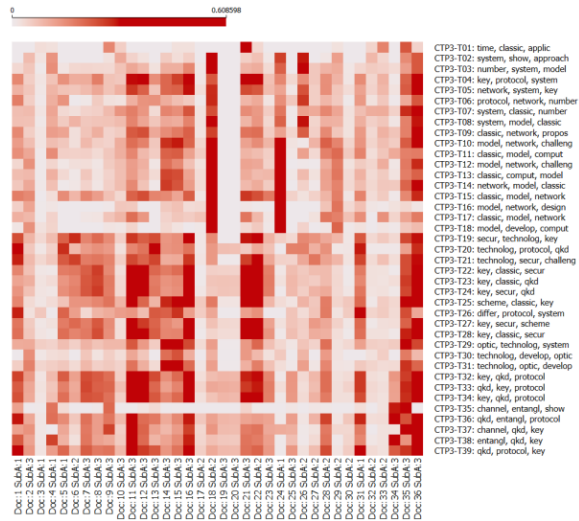

**Fig 17:** Mapping QCrypt2024 Papers to CTP3 Topics

## References

Agrawal, A., Gans, J., & Goldfarb, A. (2022). *Prediction Machines, Updated and Expanded: The Simple Economics of Artificial Intelligence*. Harvard Business Press.

Ansari, S., & Garud, R. (2009). Inter-generational transitions in socio-technical systems: The case of mobile communications. *Research Policy*, *38*(2), 382–392. https://doi.org/10.1016/j.respol.2008.11.009

Benner, M. J., & Tushman, M. L. (2015). Reflections on the 2013 Decade Award—"Exploitation, Exploration, and Process Management: The Productivity Dilemma Revisited" Ten Years Later. *Academy of Management Review*, *40*(4), 497–514. https://doi.org/10.5465/amr.2015.0042

Bennett, C. H., & Brassard, G. (2014). Quantum cryptography: Public key distribution and coin tossing. *Theoretical Computer Science*, *560*, 7–11. https://doi.org/10.1016/j.tcs.2014.05.025

Bishop, C. M. (2006). *Pattern Recognition and Machine Learning*. https://link.springer.com/book/9780387310732

Blei, D. M., & Jordan, M. I. (2006). Variational inference for Dirichlet process mixtures. *Bayesian Anal.*, *1*(1), 121–143.

Blei, D. M., & Lafferty, J. D. (2007). *A correlated topic model of science*. https://projecteuclid.org/journals/annals-of-applied-statistics/volume-1/issue-1/----Custom-HTML----A/10.1214/07-AOAS114.short

Blei, D. M., Ng, A. Y., & Jordan, M. I. (2003). Latent dirichlet allocation. *Journal of Machine Learning Research*, *3*(Jan), 993–1022.

Bogers, M., Chesbrough, H., & Moedas, C. (2018). Open Innovation: Research, Practices, and Policies. *California Management Review*, *60*(2), 5–16. https://doi.org/10.1177/0008125617745086

Cavaliere, F., Prati, E., Poti, L., Muhammad, I., & Catuogno, T. (2020). Secure Quantum Communication Technologies and Systems: From Labs to Markets. *Quantum Reports*, *2*(1), 80–106. https://doi.org/10.3390/quantum2010007

Chen, Z., Mukherjee, A., & Liu, B. (2014). Aspect extraction with automated prior knowledge learning. *Proceedings of the 52nd Annual Meeting of the Association for Computational Linguistics (Volume 1: Long Papers)*.

Dieng, A. B., Ruiz, F. J. R., & Blei, D. M. (2020). Topic Modeling in Embedding Spaces. *Transactions of the Association for Computational Linguistics*, *8*, 439–453. https://doi.org/10.1162/tacl_a_00325

Eggers, J. P., & Park, K. F. (2018). Incumbent Adaptation to Technological Change: The Past, Present, and Future of Research on Heterogeneous Incumbent Response. *Academy of Management Annals*, *12*(1), 357–389. https://doi.org/10.5465/annals.2016.0051

Floyd, S. W., & Lane, P. J. (2000). Strategizing throughout the Organization: Managing Role Conflict in Strategic Renewal. *The Academy of Management Review*, *25*(1), 154. https://doi.org/10.2307/259268

Gisin, N., Ribordy, G., Tittel, W., & Zbinden, H. (2002). Quantum cryptography. *Reviews of Modern Physics*, *74*(1), 145–195. https://doi.org/10.1103/RevModPhys.74.145

Griffiths, T. L., & Steyvers, M. (2004). Finding scientific topics. *Proc. Natl. Acad. Sci. U. S. A.*, *101 Suppl 1*(suppl_1), 5228–5235.

Gupta, A. K., Smith, K. G., & Shalley, C. E. (2006). The Interplay Between Exploration and Exploitation. *Academy of Management Journal*, *49*(4), 693–706. https://doi.org/10.5465/amj.2006.22083026

Hassija, V., Chamola, V., Saxena, V., Chanana, V., Parashari, P., Mumtaz, S., & Guizani, M. (2020). Present landscape of quantum computing. *IET Quantum Communication*, *1*(2), 42–48. https://doi.org/10.1049/iet-qtc.2020.0027

Kullback, S., & Leibler, R. A. (1951). On Information and Sufficiency. *The Annals of Mathematical Statistics*, *22*(1), 79–86.

Larochelle, H., & Lauly, S. (2012). A Neural Autoregressive Topic Model. *Advances in Neural Information Processing Systems*, *25*. https://proceedings.neurips.cc/paper/2012/hash/b495ce63ede0f4efc9eec62cb947c162-Abstract.html

Laursen, K., & Salter, A. (2006). Open for innovation: The role of openness in explaining innovation performance among U.K. manufacturing firms. *Strategic Management Journal*, *27*(2), 131–150. https://doi.org/10.1002/smj.507

Lewis, F. L., & Vrabie, D. (2009). Reinforcement learning and adaptive dynamic programming for feedback control. *IEEE Circuits and Systems Magazine*, *9*(3), 32–50. IEEE Circuits and Systems Magazine. https://doi.org/10.1109/MCAS.2009.933854

Lewis, F. L., Vrabie, D., & Vamvoudakis, K. G. (2012). Reinforcement Learning and Feedback Control: Using Natural Decision Methods to Design Optimal Adaptive Controllers. *IEEE Control Systems Magazine*, *32*(6), 76–105. IEEE Control Systems Magazine. https://doi.org/10.1109/MCS.2012.2214134

Liao, S.-K., Cai, W.-Q., Liu, W.-Y., Zhang, L., Li, Y., Ren, J.-G., Yin, J., Shen, Q., Cao, Y., Li, Z.-P., Li, F.-Z., Chen, X.-W., Sun, L.-H., Jia, J.-J., Wu, J.-C., Jiang, X.-J., Wang, J.-F., Huang, Y.-M., Wang, Q., … Pan, J.-W. (2017). Satellite-to-ground quantum key distribution. *Nature*, *549*(7670), 43–47. https://doi.org/10.1038/nature23655

Lindgreen, A., Di Benedetto, C. A., Brodie, R. J., & Jaakkola, E. (2021). How to develop great conceptual frameworks for business-to-business marketing. *Industrial Marketing Management*, *94*, A2–A10. https://doi.org/10.1016/j.indmarman.2020.04.005

Manning, C., & Schutze, H. (1999). *Foundations of Statistical Natural Language Processing*. MIT Press.

Manzalini, A. (2020). Quantum Communications in Future Networks and Services. *Quantum Reports*, *2*(1), 221–232. https://doi.org/10.3390/quantum2010014

March, J. G. (1991). Exploration and Exploitation in Organizational Learning. *Organization Science*. https://doi.org/10.1287/orsc.2.1.71

Mcauliffe, J., & Blei, D. (2007). Supervised Topic Models. *Advances in Neural Information Processing Systems*, *20*. https://proceedings.neurips.cc/paper/2007/hash/d56b9fc4b0f1be8871f5e1c40c0067e7-Abstract.html

Miao, Y., Yu, L., & Blunsom, P. (2016, June 11). Neural Variational Inference for Text Processing. *Proceedings of The 33rd International Conference on Machine Learning*. https://proceedings.mlr.press/v48/miao16.html

Mnih, V., Kavukcuoglu, K., Silver, D., Rusu, A. A., Veness, J., Bellemare, M. G., Graves, A., Riedmiller, M., Fidjeland, A. K., Ostrovski, G., Petersen, S., Beattie, C., Sadik, A., Antonoglou, I., King, H., Kumaran, D.,

Wierstra, D., Legg, S., & Hassabis, D. (2015). Human-level control through deep reinforcement learning. *Nature*, *518*(7540), 529–533. https://doi.org/10.1038/nature14236

Monarch, R. (Munro). (2021). *Human-in-the-Loop Machine Learning: Active learning and annotation for human-centered AI*. Simon and Schuster.

Mosqueira-Rey, E., Hernández-Pereira, E., Alonso-Ríos, D., Bobes-Bascarán, J., & Fernández-Leal, Á. (2022). Human-in-the-loop machine learning: A state of the art. *Artificial Intelligence Review*, *56*(4), Article 4. https://doi.org/10.1007/s10462-022-10246-w

Nazari, A., & Weiss, M. (2025). *Fine-Tuning Topics through Weighting Aspect Keywords* (arXiv:2502.08496). arXiv. https://doi.org/10.48550/arXiv.2502.08496

Ng, A. Y., Harada, D., & Russell, S. (1999). *Policy invariance under reward transformations: Theory and application to reward shaping*. In Icml (Vol. 99, pp. 278-287). https://www.teach.cs.toronto.edu/~csc2542h/fall/material/csc2542f16_reward_shaping.pdf

O'Reilly, C. A., & Tushman, M. L. (2011). Organizational Ambidexterity in Action: How Managers Explore and Exploit. *California Management Review*, *53*(4), 5–22. https://doi.org/10.1525/cmr.2011.53.4.5

O'Reilly, P. D., Rigopoulos, K., Witte, G., & Feldman, L. (2022). 2021 Cybersecurity and Privacy Annual Report. *NIST*. https://www.nist.gov/publications/2021-cybersecurity-and-privacy-annual-report

Peinelt, N., Nguyen, D., & Liakata, M. (2020). tBERT: Topic models and BERT joining forces for semantic similarity detection. *Proceedings of the 58th Annual Meeting of the Association for Computational Linguistics*, 7047–7055. https://aclanthology.org/2020.acl-main.630/

Shannon, C. E. (2001). A mathematical theory of communication. *SIGMOBILE Mob. Comput. Commun. Rev.*, *5*(1), 3–55. https://doi.org/10.1145/584091.584093

Shor, P. W. (1994). Algorithms for quantum computation: Discrete logarithms and factoring. *Proceedings 35th Annual Symposium on Foundations of Computer Science*, 124–134. https://doi.org/10.1109/SFCS.1994.365700

Srivastava, A., & Sutton, C. (2017). *Autoencoding Variational Inference For Topic Models* (arXiv:1703.01488). arXiv. https://doi.org/10.48550/arXiv.1703.01488

Sutton, R. S., & Barto, A. G. (2018). *Reinforcement Learning, second edition: An Introduction*. MIT Press.

Walrave, B., Romme, A. G. L., van Oorschot, K. E., & Langerak, F. (2017). Managerial attention to exploitation versus exploration: Toward a dynamic perspective on ambidexterity. *Industrial and Corporate Change*, *26*(6), 1145–1160. https://doi.org/10.1093/icc/dtx015

Walrave, B., van Oorschot, K. E., & Romme, A. G. L. (2011). Getting Trapped in the Suppression of Exploration: A Simulation Model: Getting Trapped in the Suppression of Exploration. *Journal of Management Studies*, *48*(8), 1727–1751. https://doi.org/10.1111/j.1467-6486.2011.01019.x

Wang, W., Guo, B., Shen, Y., Yang, H., Chen, Y., & Suo, X. (2020). Twin labeled LDA: A supervised topic model for document classification. *Applied Intelligence*, *50*, 4602–4615.

Watkins, C. J. C. H., & Dayan, P. (1992). Q-learning. *Machine Learning*, *8*(3), 279–292. https://doi.org/10.1007/BF00992698

Wu, H.-N., & Wang, M. (2024). Human-in-the-Loop Behavior Modeling via an Integral Concurrent Adaptive Inverse Reinforcement Learning. *IEEE Transactions on Neural Networks and Learning Systems*, *35*(8), 11359–11370. IEEE Transactions on Neural Networks and Learning Systems. https://doi.org/10.1109/TNNLS.2023.3259581

Xin, B., Yu, H., Qin, Y., Tang, Q., & Zhu, Z. (2020). Exploration Entropy for Reinforcement Learning. *Mathematical Problems in Engineering*, *2020*(1), 2672537. https://doi.org/10.1155/2020/2672537

Xu, J., & Durrett, G. (2018). *Spherical Latent Spaces for Stable Variational Autoencoders* (arXiv:1808.10805). arXiv. https://doi.org/10.48550/arXiv.1808.10805

Yen, G., Yang, F., & Hickey, T. (2002). Coordination of Exploration and Exploitation in a Dynamic Environment. *International Journal of Smart Engineering System Design*, *4*(3), 177–182. https://doi.org/10.1080/10255810213482

Zahra, S. A., & George, G. (2002). Absorptive Capacity: A Review, Reconceptualization, and Extension. *Academy of Management Review*, *27*(2), 185–203. https://doi.org/10.5465/amr.2002.6587995

Zhao, Q., Yang, J., Wang, Z., Chu, Y., Shan, W., & Tuhin, I. A. K. (2021). Clustering massive-categories and complex documents via graph convolutional network. In *Knowledge Science, Engineering and Management* (pp. 27–39). Springer International Publishing.